%% file: main.tex
\documentclass[runningheads]{llncs}
\usepackage{graphicx}
\usepackage{comment}

\usepackage{amsmath, amssymb, bm}
	
	\DeclareMathOperator*{\argmax}{argmax}
	\DeclareMathOperator*{\KL}{KL}
	
	\DeclareMathOperator*{\logdet}{logdet}
	\DeclareMathOperator*{\sgn}{sgn}
	\DeclareMathOperator*{\abs}{abs}
	
\usepackage{algorithm2e}
\usepackage{cite}
\usepackage{tikz}
	\usetikzlibrary{arrows,automata}
\usepackage{hyperref}
\usepackage{xr}
\usepackage{xcolor}

\usepackage{listings}
\definecolor{codegreen}{rgb}{0,0.6,0}
\definecolor{codegray}{rgb}{0.5,0.5,0.5}
\definecolor{codepurple}{rgb}{0.58,0,0.82}
\definecolor{backcolour}{rgb}{0.95,0.95,0.92}
\lstdefinestyle{mystyle}{
	backgroundcolor=\color{backcolour},   
	commentstyle=\color{codegreen},
	keywordstyle=\color{magenta},
	numberstyle=\tiny\color{codegray},
	stringstyle=\color{codepurple},
	basicstyle=\ttfamily\footnotesize,
	breakatwhitespace=false,         
	breaklines=true,                 
	captionpos=b,                    
	keepspaces=true,                 
	numbers=left,                    
	numbersep=5pt,                  
	showspaces=false,                
	showstringspaces=false,
	showtabs=false,                  
	tabsize=2
}
\begin{document}
\title{Active Learning in \\ Gaussian Process State Space Model}
%
%
\author{Hon Sum Alec Yu\inst{1, 2} \and Dingling Yao\inst{1} \and \\ Christoph Zimmer\inst{1} \and Marc Toussaint\inst{2} \and Duy Nguyen-Tuong\inst{1}}
%
\authorrunning{H.S.A. Yu et al.}

\institute{Bosch Center for Artificial Intelligence, 71272 Renningen, Germany \and Learning \& Intelligent System Lab, Technische Universit\"{a}t Berlin, Berlin, Germany}
\maketitle              
\begin{abstract}
We investigate active learning in Gaussian Process state-space models (GPSSM). Our problem is to actively steer the system through latent states by determining its inputs such that the underlying dynamics can be optimally learned by a GPSSM. In order that the most informative inputs are selected, we employ mutual information as our active learning criterion. In particular, we present two approaches for the approximation of mutual information for the GPSSM given latent states. The proposed approaches are evaluated in several physical systems where we actively learn the underlying non-linear dynamics represented by the state-space model.

\end{abstract}

\input{section_intro}
\input{section_background}

\input{section_gpssm}
\input{section_al}
\input{section_experiments}

\input{section_conclusion}

%
%
%
\bibliographystyle{splncs04}
\bibliography{reference}

\appendix
\input{supplementary}

\end{document}

%% file: section_intro.tex
\section{Introduction}\label{section_intro}

State-space models (SSMs) are a compact representation of dynamical systems as a set of input, output and state variables where the transition could be deterministic or stochastic. SSMs are widely used in practice, where applications range from aerospace-related work (e.g. \cite{faruqi2007state}), medicine (e.g. \cite{alaa2019attentive}) to economics and finance (e.g. \cite{zeng2013state}). The two important aspects of SSMs are \emph{learning} and \emph{control} and they have been a fruitful research field in classical engineering (e.g.  \cite{kirk2004optimal, friedland2012control}). Learning in SSM is about finding optimal hyperparameters so that the dynamical system is accurately captured whereas control refers to optimising a specific objective function based on inputs and states of a given SSM. The latter is often referred as \emph{optimal control}. During the last decade, both aspects of SSMs also attracted the attention from the machine learning community (e.g. \cite{frigola2015bayesian, rangapuram2018deep}). To capture the uncertainty of system dynamics, Bayesian models are often employed in the context of SSM. For example, a Gaussian Process (GP) prior \cite{rw2006gpml} is mostly used to describe the transition between states, which leads to family of Gaussian Process State-Space Models (GPSSMs) with an impressive amount of progress has been made (e.g. \cite{ko2009gp, turner2010state, frigola2013bayesian, frigola2014variational, mchutchon2015nonlinear, eleftheriadis2017identification, doerr2018probabilistic, ialongo2018closed, ialongo2019overcoming}, and more). Learning and inference of GPSSM are an active research topic in machine learning community and this paper focuses on \emph{learning} of GPSSM.

While different approaches of GPSSM mainly address the modelling issues, the question about data-efficient learning of GPSSM remains largely unanswered. Such questions go naturally to the field of Active Learning (AL), which is the process of strategically selecting and generating new data for supervised learning. The purpose of AL is to learn the proposed model with the least amount of data possible. Though the majority of recent papers in this field focused on classification problems (e.g. \cite{amin2020understanding, shui2020deep}) and rather few were for regression (e.g. \cite{seo2000gaussian, cai2016}), Schreiter et al.\cite{schreiter2015safe} applied AL in GP regression and Zimmer et al.\cite{zimmer2018safe} extended the scope to learning time-series models. Both papers employed the entropy criterion for exploration, while using a greedy selection based on the maximum predictive variance of the GP. In context of state-space models for predictive control, Buisson et al.\cite{buisson2020actively} proposed an AL approach for learning the transition dynamics via optimising the trajectory. For AL in GPSSM, Capone et al.\cite{capone2020localized} was the first to introduce the concept here. However, in both cases, Buisson et al.\cite{buisson2020actively} and Capone et al.\cite{capone2020localized} simplified the settings by assuming the states are observable and measurable. By doing so, the resulting state-space models can be actively learned in the same way as standard GP regression models in a supervised manner. 

In contrast to previous work, we present an AL strategy on GPSSM without relying on the assumption that the states are observed. That is, we actively steer the system through the latent state-space by determining its inputs such that the underlying dynamics can be optimally learned by a GPSSM. For the information criterion, we employ an approximated measure of the mutual information. We propose and discuss two different approaches for the approximation of mutual information in the context of unmeasurable, latent state-space.  

This paper is on \emph{learning} a stochastic nonlinear dynamical system actively from noisy data via probabilistic SSM. Its main contributions include:
\begin{itemize}
\item We derive tractable approximate mutual information estimates for GPSSM with latent states based on the state-of-the-art learning scheme and the technique of approximate Gaussian integral.
\item We propose an AL strategy for GPSSM with latent states based on inputs.  
\item We conduct experiments for different examples to demonstrate the usefulness of our strategy in physical systems, which is a typical application for SSM.
\end{itemize}
The reminder of this paper is organised as follow: We give a brief overview of backgrounds in section 2 and introduce GPSSM in section 3. Section 4 shows our active learning strategies and section 5 evaluates our concepts on several test scenarios. A conclusion is given in section 6. Detailed proof of propositions, experimental details and few further remarks are left in supplementary materials.

%

%% file: section_background.tex
\section{Background}\label{section_background}

AL is itself a broad topic and here we refer readers to, for example, Settles \cite{settles2009active} and Dasgupta\cite{dasgupta2011two} for an in-depth survey of the basic algorithmic and theoretical ideas. There are many different paradigms in AL. For example, \emph{Bayesian Active Learning} incorporates the Bayesian framework and is often referred to AL with GP models (e.g. \cite{houlsby2011bayesian, twomey2015bayesian}), because GPs naturally carry the uncertainty measures. Specifically, Houlsby et al.\cite{houlsby2011bayesian} applied AL in GP classification problems, where they defined an acquisition function which estimates the quantity of mutual information between model predictions and model parameters. Other variants include \emph{Batch Active Learning} and \emph{Deep Bayesian Active Learning}. While Batch AL refers to making multiple queries in parallel (e.g. \cite{kirsch2019batchbald, houlsby2011bayesian}), Deep Bayesian AL explores techniques for actively learning deep Bayesian models, e.g. Bayesian convolutional neural network\cite{gal2017deep}. For AL with regression models, Seo et al.\cite{seo2000gaussian} proposed a test point rejection strategy based on the posterior variance. Krause et al.\cite{krause2007nonmyopic} proposed an optimal exploration strategy for GP regression models, while providing bounds on the advantage of using AL. Srinivas et al.\cite{srinivas2012information} further showed convergence rate for AL in GPs under specific kernels. Schreiter et al.\cite{schreiter2015safe} extended AL in GP regression by introducing constraints and Zimmer et al.\cite{zimmer2018safe} extended constrained AL in GP regression to time-series. They proposed using the determinant of the predictive covariance matrix defined in the GP as optimality criterion along with theoretical guarantees.

In dynamics modelling, SSM has been a long-lasting topic with enormous literature across different disciplines (e.g. \cite{kirk2004optimal, friedland2012control, zeng2013state}). In machine learning, the combination between SSM and GP began arguably from Wang et al.\cite{wang2006gaussian}, where they learned the latent state of GPSSM via maximum a posteriori. Turner et al.\cite{turner2010state} extended it to learning the transition function. Frigola et al.\cite{frigola2013bayesian} presented a Bayesian treatment based on particle Markov-Chain-Monte-Carlo and Frigola et al.\cite{frigola2014variational} further introduced a variational inference scheme to overcome the computational complexity of Monte-Carlo based methods. This becomes the state-of-the-art and due to the fact that this approach depends heavily on the approximated function, most recent work introduced their own variational inference scheme with their own advantages and disadvantages (e.g. \cite{mchutchon2015nonlinear, doerr2018probabilistic, ialongo2018closed, ialongo2019overcoming}). For example, Doerr et al.\cite{doerr2018probabilistic} and Ialongo et al.\cite{ialongo2019overcoming} introduced their specific variational inference scheme to account for the dependence between transition function and latent states, which were treated as independent in earlier work.

To the best of our knowledge, AL and GPSSM have not been discussed together until Buisson et al.\cite{buisson2020actively} first brought the topic of AL in learning the transition function using GP. They stated that the dynamical problem, i.e. learning the transition function, is fundamentally different from a static AL in GP problem, because we need to steer the system to certain states through the unknown transition function with a sequence of actions, which is the only component we can control. For each round of exploration, the next input is actively picked so that the model error can be maximally reduced. They proposed their AL strategy using the maximum entropy. Another recent work on AL in GPSSM is by Capone et al.\cite{capone2020localized}. They proposed taking the most informative data point while employing the mutual information between the most informative point and reference points. The states, in the existing work on AL in GPSSM, are hitherto assumed to be observable such that the model can be trained in a fashion similar to GP regression. Namely, in Capone et al.\cite{capone2020localized}, the fact that states are observable simplify the estimation of the mutual information, resulting in two optimisation problems to solve in parallel. 

Our paper differs from the previous work by presenting an AL strategy on GPSSM with states that are \emph{latent}, resulting in a different estimation of the mutual information as exploration criterion while extending the scope of AL for GPSSM. We leave further quantitative discussion in section \ref{section_al}.

%% file: section_gpssm.tex
\section{Gaussian Process State-Space Model with Inputs} \label{section_gpssm_c}

We consider a discrete-time sequence of $T$ observations $\bm{y}_{1: T} \!\equiv\! \lbrace \bm{y}_t \rbrace^{T}_{t = 1}$, where each observed point $\bm{y}_t \!\in\! \mathcal{Y} \!\subseteq\! \mathbb{R}^{d_y}$ is generated by a corresponding latent variable $\bm{x}_t \!\in\! \mathcal{X} \!\subseteq\! \mathbb{R}^{d_x}$ and the previous step control $\bm{c}_{t-1} \!\in\! \mathcal{C} \!\subseteq\! \mathbb{R}^{d_c}$. The collection of latent states and controls are denoted by $\bm{x}_{0: T} \!\equiv\! \lbrace \bm{x}_t \rbrace^{T}_{t = 0}$ and $\bm{c}_{0: T} \!\equiv\! \lbrace \bm{c}_t \rbrace^{T}_{t = 0}$, respectively. These latent variables and controls are assumed to satisfy the Markov property, meaning that any $\bm{x}_{t+1}$ can be generated by only conditioning on $\bm{x}_t$, $\bm{c}_t$ and the transition function $f: \mathbb{R}^{d_x} \!\times\! \mathbb{R}^{d_c} \!\rightarrow\! \mathbb{R}^{d_x}$. To align with previous work, we use Gaussian distribution for both transition and observation density function. We also simplify the mean of the observation function by using linear mapping and tackle multivariate latent states by placing a GP prior on each dimension separately, as in, for instance, Ialongo et al.\cite{ialongo2019overcoming}. 

GPSSM is defined to be a probabilistic SSM with a GP prior over the transition function $f$, specified by 
\begin{align}
f &\sim \mathcal{GP}(m(\cdot), k(\cdot, \cdot)),  \label{gpssm_equation1} \\
\bm{x}_0 &\sim p(\bm{x}_0), \label{gpssm_equation2} \\
\bm{x}_t \vert f(\bm{x}_{t-1}, \bm{c}_{t-1}) &\sim \mathcal{N}(\bm{x}_t \vert f(\bm{x}_{t-1}, \bm{c}_{t-1}), \bm{Q}),  \label{gpssm_equation3} \\
\bm{y}_t \vert \bm{x}_t &\sim \mathcal{N}(\bm{y}_t \vert \bm{Cx}_t + \bm{d}, \bm{R})  \label{gpssm_equation4},
\end{align}
where in equation \ref{gpssm_equation1}, $f$ is a GP governed by a given mean function $m(\cdot)$ and positive definite covariance function $k(\cdot, \cdot)$. The initial state $p(\bm{x}_0)\!=\!\mathcal{N}(\bm{x}_0 \vert \bm{\mu}_0, \bm{\Sigma}_0)$ in equation \ref{gpssm_equation2} is assumed known. $\bm{C}$ and $\bm{d}$ are parameters in the linear mapping, whereas $\bm{Q}$ and $\bm{R}$ are covariance matrices which capture process and observation noise, respectively. The advantages of linear mapping are that linear mean won’t limit the range of systems that can be modelled and this reduces the non-identifiabilities between transitions and emissions. As a convention from previous work, we denote $\bm{f}_t \!\equiv\! f(\bm{x}_{t-1}, \bm{c}_{t-1})$ and equation \ref{gpssm_equation3} becomes 
\begin{equation}\label{gpssm_equation3_2} 
\bm{x}_t \vert \bm{f}_t \sim \mathcal{N}(\bm{x}_t \vert \bm{f}_t, \bm{Q}).
\end{equation}
For brevity, we write $\tilde{\bm{x}}_\ast \!=\! (\bm{x}_\ast, \bm{c}_\ast)$, where $\ast$ could be a specific index or a collection of indices. Thus, the matrix of covariance functions are denoted by $K_{i: j} \!:=\! (k(\tilde{\bm{x}}_s, \tilde{\bm{x}}_t))^j_{s, t=i}$. For convenience, we write $k(\tilde{\bm{x}}_{i:j}, \tilde{\bm{x}}_k) \!\equiv\! \left( k(\tilde{\bm{x}}_i, \tilde{\bm{x}}_k), \cdots, k(\tilde{\bm{x}}_j, \tilde{\bm{x}}_k) \right)$ as a vector collection of kernel evaluations. It holds that $k(\tilde{\bm{x}}_{i:j}, \tilde{\bm{x}}_k)^T \!=\! k(\tilde{\bm{x}}_k, \tilde{\bm{x}}_{i:j})$. 

\begin{figure}[t!]
	\centering
	\resizebox{.75\textwidth}{!}{
	\begin{tikzpicture}
	[->, >=stealth', shorten >=1pt, auto, node distance=1.8cm,
	semithick]
	\tikzstyle{every state}=[circle, text=black]
	
	\node[state, fill=green!40, scale=1.2, inner sep=0pt]	(x0)					{$\bm{x}_0$};
	\node[state, scale=1.2, inner sep=0pt]			(x1)		[right of=x0]			{$\bm{x}_1$};
	\node[state, draw=none]	(xdots)		[right of=x1]			{$\cdots$};
	\node[state, scale=1.2, inner sep=0pt]			(xtMinus1)	[right of=xdots] 		{$\bm{x}_{t-1}$};
	\node[state, scale=1.2, inner sep=0pt]			(xt)		[right of=xtMinus1]		{$\bm{x}_t$};
	\node[state, scale=1.2, inner sep=0pt]			(xtPlus1)	[right of=xt]			{$\bm{x}_{t+1}$};
	\node[state, draw=none]	(xtPlusN)	[right of=xtPlus1]		{$\cdots$};
	\node[state, fill=gray!25, scale=1.2, inner sep=0pt]			(f1)		[above of=x1]			{$\bm{f}_1$};
	\node[state, draw=none]	(fdots)		[right of=f1]			{$\cdots$};
	\node[state, fill=gray!25, scale=1.2, inner sep=0pt]			(ftMinus1)	[right of=fdots] 		{$\bm{f}_{t-1}$};
	\node[state, fill=gray!25, scale=1.2, inner sep=0pt]			(ft)		[right of=ftMinus1]		{$\bm{f}_t$};
	\node[state, fill=gray!25, scale=1.2, inner sep=0pt]			(ftPlus1)	[right of=ft]			{$\bm{f}_{t+1}$};
	\node[state, draw=none]	(ftPlusN)	[right of=ftPlus1]		{$\cdots$};
	\node[state, fill=yellow!40, scale=1.2, inner sep=0pt]	(c0)	[above left of=f1]		{$\bm{c}_0$};
	\node[state, draw=none]	(cdots)		[right of=c0]			{$\cdots$};
	\node[state, draw=none]	(cdots2)	[right of=cdots] 		{$\cdots$};
	\node[state, fill=yellow!40, scale=1.2, inner sep=0pt]	(ctMinus1)	[right of=cdots2]	{$\bm{c}_{t-1}$};
	\node[state, fill=yellow!40, scale=1.2, inner sep=0pt]	(ct)		[right of=ctMinus1]	{$\bm{c}_t$};
	\node[state, fill=green!40, scale=1.2, inner sep=0pt]	(y1)	[below of=x1]			{$\bm{y}_1$};
	\node[state, draw=none]	(ydots)		[below of=xdots]		{$\cdots$};
	\node[state, fill=green!40, scale=1.2, inner sep=0pt]	(ytMinus1)	[below of=xtMinus1]	{$\bm{y}_{t-1}$};
	\node[state, fill=green!40, scale=1.2, inner sep=0pt]	(yt)		[below of=xt]		{$\bm{y}_t$};
	\node[state, fill=green!40, scale=1.2, inner sep=0pt]	(ytPlus1)	[below of=xtPlus1]	{$\bm{y}_{t+1}$};
	
	\path	(x0)		edge					node	{}	(f1)
			(c0) 		edge					node	{}	(f1)
			(f1)		edge[-, line width=3.5pt]	node	{}	(fdots)
			(fdots) 	edge[-, line width=3.5pt]	node	{}	(ftMinus1)
			(ftMinus1)	edge[-, line width=3.5pt]	node	{}	(ft)
			(ft)		edge[-, line width=3.5pt]	node	{}	(ftPlus1)
			(ftPlus1)	edge[-, line width=3.5pt]	node	{}	(ftPlusN)
			(x1)		edge					node	{}	(fdots)
			(xdots)		edge					node	{}	(ftMinus1)
			(cdots2)	edge					node	{}	(ftMinus1)
			(ctMinus1)	edge					node	{}	(ft)
			(ct)		edge					node	{}	(ftPlus1)
			(xtMinus1)	edge					node	{}	(ft)
			(xt)		edge					node	{}	(ftPlus1)
			(xtPlus1)	edge					node	{}	(ftPlusN)
			(f1)		edge					node	{}	(x1)
			(x1)		edge					node	{}	(y1)
			(ftMinus1)	edge					node	{}	(xtMinus1)
			(xtMinus1)	edge					node	{}	(ytMinus1)
			(ft)		edge					node	{}	(xt)
			(xt)		edge					node	{}	(yt)
			(ftPlus1)	edge					node	{}	(xtPlus1)
			(xtPlus1)	edge					node	{}	(ytPlus1)
	;
	\end{tikzpicture}}
\caption{Graphical model of GPSSM with control inputs as defined in equation \ref{gpssm_equation1} - \ref{gpssm_equation4}. The observations $\bm{y}_t$ and initial state $\bm{x}_0$ are assumed known (filled in green). Control inputs $\bm{c}_t$ are independent variables (filled in yellow) whereas latent variables are denoted by $\bm{x}_t$. We use a thick, straight line to connect $\bm{f}_t \!\equiv\! f(\bm{x}_{t-1}, \bm{c}_{t-1})$ (filled in gray) to show that all variables are fully connected.}
\label{GPSSM_GraphicalModel}
\end{figure}
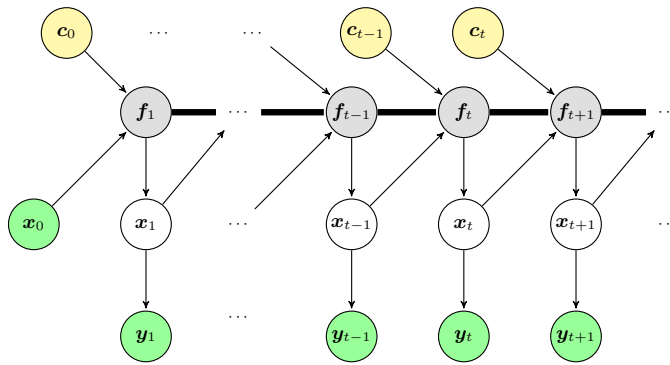

The control inputs $\bm{c}_t$ can be viewed as an augmented latent state. Since $\bm{c}_t$ is not a random variable, we always condition on this quantity. Most previous papers related to this model were either studied without controls or, if a control is involved, by assuming that this quantity was randomly distributed \cite{mattos2016latent}. We, however, construct an AL strategy based on $\bm{c}_t$ to learn the GPSSM efficiently. Hence, we purposely keep this term in our definitions. Figure \ref{GPSSM_GraphicalModel} shows a graphical model of such a setting. 
We would like to readdress here that $\bm{c}_t$ is a known quantity where our task is to optimise the value to select. This should not be confused with control in other contexts such as that in RL.

To complete the structure of GPSSM, let us define  $\bm{f}_{1:T} \!\equiv\! \lbrace \bm{f}_t \rbrace^T_{t=1}$ and we can write the joint density as follows (e.g. chapter 3 in Frigola\cite{frigola2015bayesian}),
\begin{equation}\label{gpssm_jd}
p(\bm{y}_{1:T}, \bm{x}_{0:T}, \bm{f}_{1:T}) = p(\bm{x}_0) \prod^T_{t=1} p(\bm{y}_t \vert \bm{x}_t) p(\bm{x}_t \vert \bm{f}_t) p(\bm{f}_t \vert \bm{x}_{0:t-1}, \bm{f}_{1:t-1}).
\end{equation}
In particular, from equation \ref{gpssm_equation1}, the last factor in equation \ref{gpssm_jd} is derived in the same manner as the posterior in GP regression, given by
\begin{equation}\label{gpssm_f_cond_prev}
p(\bm{f}_t \vert \bm{x}_{0:t-1}, \bm{f}_{1:t-1}) = \mathcal{N}(\bm{f}_t \vert \mathcal{M}_{t-1}, \mathcal{K}_{t-1}),
\vspace{-0.1cm} 
\end{equation}
where
\begin{equation*}
\begin{aligned}
\mathcal{M}_{t-1} &= m(\tilde{\bm{x}}_{t-1}) + k(\tilde{\bm{x}}_{t-1}, \tilde{\bm{x}}_{0: t-2}) K^{-1}_{0: t-2} (\bm{f}_{1: t-1} - m(\tilde{\bm{x}}_{0: t-2}))^T, \\
\mathcal{K}_{t-1} &= k(\tilde{\bm{x}}_{t-1}, \tilde{\bm{x}}_{t-1}) - k(\tilde{\bm{x}}_{t-1}, \tilde{\bm{x}}_{0: t-2}) K^{-1}_{0: t-2}  k(\tilde{\bm{x}}_{0: t-2}, \tilde{\bm{x}}_{t-1}),
\end{aligned}
\end{equation*}
and the term $\bm{f}_{1: t-1} \!-\! m(\tilde{\bm{x}}_{0: t-2})$ is written as 
\begin{equation*}
\bm{f}_{1: t-1} - m(\tilde{\bm{x}}_{0: t-2}) \!\equiv\! (\bm{f}_1 - m(\tilde{\bm{x}}_0), \cdots, \bm{f}_{t-1} - m(\tilde{\bm{x}}_{t-2})).
\end{equation*}

\subsection{Learning and Prediction in GPSSM}

In learning the posterior, our target is to compute
\[ p(\bm{x}_{0:t}, \bm{f}_{1:t} \vert \bm{y}_{1:t}) = p(\bm{x}_{0:t}, \bm{f}_{1:t} , \bm{y}_{1:t}) / p(\bm{y}_{1:t}). \]

The major challenge is the computation of $p(\bm{y}_{1:t})$. Throughout this paper, we are going to apply \emph{variational inference} to train GPSSM, which is a technique based on making assumptions about the posterior over latent variables that leads to a tractable lower bound, often called \emph{evidence lower bound} (ELBO). There are two reasons for choosing this approach, which is currently state-of-the-art in training GPSSM. First, variational inference is computationally efficient. Second, this approach will lead to an approximation of the posterior, which can be carried to the AL strategy, as we will present further in section \ref{section_al}. 

The first step of using variational inference in GPSSM is to introduce $M$ ($M\ll T$) inducing points $\bm{u} \!=\! \bm{u}_{1:M} \!=\! \lbrace \bm{u}_i \rbrace^M_{i=1}$ with their corresponding inputs. This is referred as sparse GP technique (see e.g. \cite{titsias2009variational, matthews2016sparse}) and the joint density is
\begin{equation}\label{gpssm_jd_u}
p(\bm{y}_{1:T}, \bm{x}_{0:T}, \bm{f}_{1:T}, \bm{u}) = p(\bm{x}_0) p(\bm{u}) \prod^T_{t=1} p(\bm{y}_t \vert \bm{x}_t) p(\bm{x}_t \vert \bm{f}_t)p(\bm{f}_t \vert \bm{x}_{0:t-1}, \bm{f}_{1:t-1}, \bm{u}).
\end{equation}
Then, the ELBO $\mathcal{L}_t$ to the log marginal likelihood of $p(\bm{y}_{1:t})$ in equation \ref{gpssm_jd}, which is based on KL divergence (see e.g. chapter 8.5 in Cover\cite{cover2006elements}), is given by \[ \log(p(\bm{y}_{1:t})) \!=\! \mathcal{L}_t \!+\! \KL\left[q(\bm{x}_{0:t}, \bm{f}_{1:t}) \| p(\bm{x}_{0:t}, \bm{f}_{1:t} \vert \bm{y}_{1:t}) \right] \] for $t\!=\!1,\!\cdots\!, T$, where $q(\cdot)$ is the distribution function to approximate $p(\cdot)$. With further derivations from variational inference methodology, $\mathcal{L}_t$ is given by 
\begin{equation}\label{gpssm_elbo}
\mathcal{L}_t = \int q(\bm{x}_{0:t}, \bm{f}_{1:t}, \bm{u}) \log\left( \frac{p(\bm{y}_{1:t}, \bm{x}_{0:t}, \bm{f}_{1:t}, \bm{u})}{q(\bm{x}_{0:t}, \bm{f}_{1:t}, \bm{u})} \right) d\bm{x}_{0:t} d\bm{f}_{1:t} d\bm{u}.
\end{equation}
Analogously to Ialongo et al.\cite{ialongo2019overcoming}, we set
\begin{equation}\label{q_func}
q(\bm{x}_{0:t}, \bm{f}_{1:t}, \bm{u}) = q(\bm{u}) q(\bm{x}_0) \prod^t_{i=1} q(\bm{x}_i \vert \bm{f}_i )  p(\bm{f}_i \vert \bm{f}_{1:i-1}, \bm{x}_{0:i-1}, \bm{u})
\end{equation}
which leads to equation 17 of \cite{ialongo2019overcoming}
\begin{equation} \label{gpssm_elbo_2}
\begin{aligned}
\mathcal{L}_T = & \int \sum^T_{t=1} q(\bm{x}_{0:T}) \log\left( p(\bm{y}_t \vert \bm{x}_t) \right) d\bm{x}_{0:T} - \sum^T_{t=1} \int q(\bm{f}_t) \KL\left[ q(\bm{x}_t \vert \bm{f}_t) \| p(\bm{x}_t \vert \bm{f}_t) \right] d\bm{f}_t \\
&- \KL\left[ q(\bm{x}_0) \| p(\bm{x}_0) \right] - \KL\left[ q(\bm{u}) \| p(\bm{u}) \right].
\end{aligned}
\end{equation}
Under sparse GP approximation, we can specify a free Gaussian density on the function values $\bm{u}$ giving $q(\bm{u}) \!=\! \mathcal{N}(\bm{u} \vert \bm{\mu}_{\bm{u}}, \bm{\Sigma}_{\bm{u}})$. The only term that requires further specification is $q(\bm{x}_i \vert \bm{f}_i)$ and \cite{ialongo2019overcoming} presented $q(\bm{x}_i \vert \bm{f}_i) \!=\! \mathcal{N}(\bm{x}_i \vert \bm{A}_{i-1} \tilde{\bm{f}}_{i-1} + \bm{b}_{i-1}, \bm{S}_{i-1})$, where $\tilde{\bm{f}}_i$ is a free variational parameter. $\bm{A}_i$, $\bm{b}_i$ and $\bm{S}_i$ are other free variational parameters depending on the choice of $\tilde{\bm{f}}_i$. This generalises a number of previous work. For example, by setting $\bm{f}_i \!=\! \bm{x}_i$ in $q(\cdot \vert \cdot)$ and $q(\bm{x}) \!=\! q(\bm{x}_0)$, we recover the expression from \cite{frigola2014variational} and the optimal $q(\bm{x})$ can be solved by calculus of variations. If we set $\tilde{\bm{f}}_i \!=\! \bm{x}_i$, then we have the Gaussian factorised approximation of \cite{mchutchon2015nonlinear}. In \cite{ialongo2019overcoming}, they proposed setting $\tilde{\bm{f}}_i = k(\bm{x}_i, \bm{u}) K^{-1}_{\bm{uu}} \bm{u}$ and the training turned out to be marginally better if a sufficiently large amount of iterations are allowed.

We choose to set $\tilde{\bm{f}}_i \!=\! \bm{x}_i$  throughout this paper, which is the approach by \cite{mchutchon2015nonlinear}. This is because their approach has shown to have a stable improvement in model accuracy during the training phase. Also, its linear factorising nature, given a fixed and often small number of allowed iterations, is also beneficial for training time. Such setting leads to $q(\bm{x}_i \vert \bm{f}_i) \!=\! \mathcal{N}(\bm{x}_i \vert \bm{A}_{i-1} \bm{x}_{i-1} + \bm{b}_{i-1}, \bm{S}_{i-1})$, with the free parameters $\bm{S}_{i-1} \!=\! (\bm{Q}^{-1} + \bm{C}^T\bm{R}^{-1}\bm{C})^{-1}$, $\bm{A}_{i-1} \!=\! \bm{S}_{i-1}\bm{Q}^{-1}$ and $\bm{b}_{i-1} \!=\! \bm{S}_{i-1}\bm{C}^T\bm{R}^{-1}(\bm{y}_t - \bm{d})$. 
%
%

The prediction with GPSSM can be done rather cheaply.  Suppose we would like to predict the new observation $\bm{y}^\star$ based on a new control input $\bm{c}^\star$, by defining all training data as $\mathcal{D} \!\equiv\! \lbrace \bm{y}_{1:T}, \bm{x}_{0:T} \rbrace$, we can show easily \cite{frigola2014variational} that the predictive distribution $p(\bm{y}^\star \vert \bm{c}^\star, \mathcal{D})$ is given by
\begin{equation}\label{gpssm_pred}
p(\bm{y}^\star \vert \bm{c}^\star, \mathcal{D}) \!=\! \mathcal{N}(\bm{y}^\star \vert \bm{C} f(\bm{x}_T, \bm{c}^\star) \!+\! \bm{d}, \bm{R} \!+\! \bm{CQC}^T).
\end{equation}
In practice, we sample $f(\bm{x}_T, \bm{c}^\star)$, where the latent states $\bm{x}_{1:T}$ are already estimated via training. Then, we predict the new observation based on drawing samples from the distribution. This is the approach used by \cite{ialongo2019overcoming}.

%% file: section_al.tex
\section{Active Learning Strategies}
\label{section_al}

The previous section presents the modelling of GPSSM and in this section, we are interested in acquiring a strategy to actively learn the GPSSM by steering the system through the latent state-space. We employ AL as a query strategy to pick the most informative new control input $\bm{c}^\ast_t$ while observing $\bm{y}_t$, and to learn the model as we explore. Contrary to the two previous work \cite{buisson2020actively, capone2020localized}, we treat the states as latent and unknown and $\bm{c}_t$ is the only controllable variable.  
For the exploration criterion, we use approximated mutual information. The usage of mutual information for AL has been well motivated by, for example, Krause et al.\cite{krause2008near}. They pointed out that the mutual information might lead to a more accurate model than the differential entropy. This quantity was also shown to be the same as minimising the expected uncertainty of the model\cite{ertin2003maximum}.

First, let us recall from chapter 8.5 of Cover\cite{cover2006elements} that the mutual information between any two sets of random variables $\bm{Y}$ and $\bm{F}$ with joint density $p(\bm{Y}, \bm{F})$ is defined as
\begin{equation*}
I(\bm{Y}; \bm{F}) = \int p(\bm{Y}, \bm{F}) \log\left(\frac{p(\bm{Y}, \bm{F})}{p(\bm{Y})p(\bm{F})}\right) d\bm{Y} d\bm{F}
\end{equation*}
and a well-known relationship between mutual information and differential entropy $h(\cdot)$ is given by $I(\bm{Y}; \bm{F}) \!=\! h(\bm{Y}) \!-\! h(\bm{Y} \vert \bm{F})$.

For steering the system through latent state-space while gathering information for efficiently learning the model, a sensible quantity for exploration is the mutual information between the latest observations $\bm{y}_{t+1}$ and latest predicted transition functions $\bm{f}_{t+1}$. This quantity is an extension to the active strategy from Buisson et al.\cite{buisson2020actively} by using mutual information as a criterion. That is,                                  
\begin{equation}\label{opt_c_MI_latest}      
\bm{c}^\ast_t = \argmax_{\bm{c}_t \in \mathcal{C}} I(\hat{\bm{y}}_{t+1}; \bm{f}_{t+1}),
\end{equation}
where $\bm{c}_t$ is placed within the term $\bm{f}_{t+1} \!\equiv\! f(\bm{x}_t, \bm{c}_t)$. We use $\hat{\bm{y}}_{t+1}$ because this term is provided from predicting GPSSM with a given new control $\bm{c}_t$. Altogether, $\bm{c}_t$ is the only independent variable. 

This expression is intractable and there have been enormous efforts in estimating it (e.g. \cite{perez2009estimation, lombardi2016nonparametric, hoffman2016elbo, alemi2018fixing, poole2019variational}). However, most previous work assumed that we have little or no information on both entropies in the mutual information expression and this comes down to the fact that we do not know the corresponding probability density function (pdf). Their approach includes deriving bounds but extra functions such as \emph{critic} or \emph{variational parameters} are often required (e.g. section 2 in \cite{poole2019variational}). However, should we have some information about this quantity, it is much better to make use of this known information instead. Therefore, we derive our estimate for the mutual information by directly approximating the pdf. The main advantages of our method versus others are: (1) As discussed in the previous paragraph, other methods often require extra functions called \emph{critic} or \emph{variational parameters}, or some free parameters to set. Finding the best settings of these extra functions and parameters could be complicated at times.  Mcallester et al.\cite{mcallester2020formal} also showed that there are statistical limitations in various bounds of mutual information and a better approach is to measure this quantity as a difference in entropies. (2) Our mathematical derivation is purely via inequalities and numerical approximations, resulting an approximated closed form. Hence, the final expression derived can be easily implemented.

The quantity $I(\bm{y}_t; \bm{f}_t)$ is often studied in SSM (e.g. \cite{ertin2003maximum}), motivated by the assumption of Markov property. However, if the Markov assumption is not completely valid, as there are (weak) dependencies between the states, the mutual information between all observations $\bm{y}_{1: t+1}$ and all predictions from the transition function $\bm{f}_{1: t+1}$ within a time-horizon might present a better quantity. To the best of our knowledge, we are the first to propose using mutual information between all variables in time, and the AL strategy presented in equation \ref{opt_c_MI_latest} can be viewed as a special case of this general formulation. Thus, another more general AL strategy is
\begin{equation}\label{opt_c_MI_all}
\bm{c}^\ast_t = \argmax_{\bm{c}_t \in \mathcal{C}} I(\bm{y}_{1:t}, \hat{\bm{y}}_{t+1}; \bm{f}_{1:t+1}).
\end{equation}
Again, the term $\hat{\bm{y}}_{t+1}$ is provided from predicting GPSSM. This also leads to the third advantage of our method: (3) To the best of our knowledge, we are the first to estimate the total mutual information by leveraging the inference scheme of the model. This turns out to be computationally efficient as well. 

For clarity, we refer $I(\hat{\bm{y}}_{t+1}; \bm{f}_{t+1})$ as \emph{latest} mutual information (latMI) and $I(\bm{y}_{1:t}, \hat{\bm{y}}_{t+1}; \bm{f}_{1:t+1})$ as \emph{total} mutual information (totMI). In the remainder of this section, we present how the two quantities can be approximated for GPSSM with latent states. We first present an estimation scheme for the \emph{latest} mutual information to align with previous work. Then, we proceed to the estimation of our proposed \emph{total} mutual information. Given the estimates of mutual information, we can formulate the AL strategy for GPSSM, as summarised in algorithm \ref{algoALGPSSM}.
\begin{algorithm}[t]
	\SetKwData{Left}{left}
	\SetKwData{This}{this}
	\SetKwData{Up}{up}
	\SetKwFunction{Union}{Union}
	\SetKwFunction{FindCompress}{FindCompress}
	\SetKwInOut{Input}{input}\SetKwInOut{Output}{output}
	
	\Input{ Initial $T$ observations $\bm{y}_{1: T}$ and control inputs $\bm{c}_{0: T}$. Initial state $\bm{x}_0$.}
	\Output{An optimised GPSSM after taking $N$ exploration steps, the final data set including the explored observations $\bm{y}_{1: T+N}$ and the corresponding controls $\bm{c}_{0: T+N}$.}
	\BlankLine
	\emph{Train the initial GPSSM}
	\For{$t = T, T+1, \cdots, T+N$}
	{{1. Solve $\bm{c}^\ast_t$ via equation (\ref{opt_c_MI_all}) (or (\ref{opt_c_MI_latest}))}\;
		{2. Evaluate $\bm{y}_{t+1}$ from the system}\;
		{3. Update the training data set}\;
		{4. Retrain GPSSM to update hyperparameters}\;
		{5. $t = t + 1$}\;	
	}
	\caption{\emph{Active Learning in GPSSM} on control through approximation of mutual information}
	\label{algoALGPSSM}
\end{algorithm}

\subsection{Computation of latest Mutual Information $I(\bm{y}_t; \bm{f}_t)$}                                                                                                                                                                             
To estimate the \emph{latest} mutual information $I(\bm{y}_t; \bm{f}_t)$, we employ the Gaussian approximate integral derived by Girard\cite{girard2004approximate}. Here, for two random variables $\bm{x}$ and $\bm{y}$ with $p(\bm{y} \vert \bm{x}) \!=\! \mathcal{N}(\bm{y} \vert \mu(\bm{x}), \sigma^2(\bm{x}))$ and $p(\bm{x} \vert \bm{u}, \bm{\Sigma}_{\bm{x}}) \!=\! \mathcal{N}(\bm{x} \vert \bm{u}, \bm{\Sigma}_{\bm{x}})$ for some mean $\bm{u}$ and variance $\Sigma_{\bm{x}}$, the Gaussian approximation yields
\begin{equation*}
\int p(\bm{y} \vert \bm{x}) p(\bm{x} \vert \bm{u}, \bm{\Sigma}_{\bm{x}}) d\bm{x} \approx \mathcal{N}(M(\bm{u}, \bm{\Sigma}_{\bm{x}}), V(\bm{u}, \bm{\Sigma}_{\bm{x}})),
\vspace{-0.1cm}                                           \end{equation*}
where $M(\bm{u}, \bm{\Sigma}_{\bm{x}})$ and $V(\bm{u}, \bm{\Sigma}_{\bm{x}})$ are integral functions to be evaluated (see supplementary materials for details). Based on this approximation, $I(\bm{y}_t; \bm{f}_t)$ can be approximately computed as in the following proposition.
\begin{proposition}
Given the definition of GPSSM from equation \ref{gpssm_equation1} - \ref{gpssm_equation4}, as well as the notation of $\mathcal{M}_t$ and $\mathcal{K}_t$ defined in equation \ref{gpssm_f_cond_prev}. We define the approximation of the following integrals as
\begin{equation*}
\begin{aligned}
\mathcal{N}(\bm{f}_1 \vert M_1, V_1) &:= \mathcal{N}(\bm{f}_1 \vert M(\bm{\mu}_0, \bm{\Sigma}_0), V(\bm{\mu}_0, \bm{\Sigma}_0)) \\
&\approx \int \mathcal{N}(\bm{f}_1 \vert \mathcal{M}_0, \mathcal{K}_0) \mathcal{N}(\bm{x}_0 \vert \bm{\mu}_0, \bm{\Sigma}_0) d\bm{x}_0
\end{aligned}
\end{equation*}
and, recursively, for all $t = T, T+1, \cdots, T+N$, 
\begin{equation*}
\mathcal{N}(\bm{f}_t \vert M_t, V_t) \approx \int \mathcal{N}(\bm{f}_t \vert \mathcal{M}_{t-1}, \mathcal{K}_{t-1}) \mathcal{N}(\bm{x}_{t-1} \vert M_{t-1}, V_{t-1} + \bm{Q}) d \bm{x}_{t-1}.
\end{equation*}
Then, the latest mutual information is approximately
\begin{equation}\label{eq_prop_mi_latest}
I(\bm{y}_t; \bm{f}_t) \approx \frac{1}{2} \log\left(\frac{\det(\bm{R} + \bm{C}(V_t + \bm{Q})\bm{C}^T)}{\det(\bm{R} + \bm{CQC}^T)}\right).
\end{equation}
\end{proposition} 
\begin{proof}[Sketch]
Based on the definition of mutual information, the numerator in equation \ref{eq_prop_mi_latest} relies on the approximation via \emph{moment matching}, whereas the denominator in equation \ref{eq_prop_mi_latest} can be computed directly. For more details, we refer the interested reader to the supplementary materials.
\end{proof}

\subsection{Computation of total Mutual Information $I(\bm{y}_{1:t}; \bm{f}_{1:t})$}

The ELBO $\mathcal{L}_t$ given by equation \ref{gpssm_elbo_2} can be utilised to estimate the \emph{total} mutual information. Since this quantity is estimated via samples, this quantity is sample driven and the following proposition presents our computational approach.
\begin{proposition}\label{prop_mi_total}
Given the definition of GPSSM from equation \ref{gpssm_equation1} - \ref{gpssm_equation4}, as well as the expression of the ELBO $\mathcal{L}_t$ from equation \ref{gpssm_elbo_2}, for $t = T, T+1, \cdots, T+N$, if $S$ samples are drawn, the $s$-th sample estimate ($s = 1, \cdots, S$) of the mutual information between observations $\bm{y}_{1:t}$ and the prediction $\bm{f}_{1:t}$, denoted by $i_s$, is bounded by
\begin{equation}\label{eq_prop_totalMI_persample}
i_s \leq  \sum^t_{i=1} \log(\mathcal{N}_s(\bm{y}_i \vert \bm{Cf}_i + \bm{d}, \bm{R} + \bm{CQC}^T)) - \mathcal{L}_{t, s},
\vspace{-0.1cm}
\end{equation}
where $\mathcal{N}_s(\cdot)$ and $\mathcal{L}_{t, s}$ are the $s$-th sample estimate of the normal distribution and ELBO $\mathcal{L}_t$, respectively. 
The total mutual information is then approximated by $I(\bm{y}_{1:t}; \bm{f}_{1:t}) \approx \frac{1}{S}\sum^S_{s=1} i_s$.
\end{proposition}

\begin{proof}[Sketch]
The proof employs mainly the definition of mutual information. The intractable part is bounded via ELBO, and the other term can be computed directly. Since it is a sample based approach, the bound from a sample becomes an approximation after taking all samples. For more details, we refer the interested reader to the supplementary materials.
\end{proof}
\begin{remark}
In computing equation \ref{eq_prop_totalMI_persample}, the sampling nature comes from the specific value of $\bm{x}_0$ drawn, and this value will be different between samples. Therefore, the evaluated $i_s$ will also be different.

The efficiency of this approach heavily relies on the inference of GPSSM and it is possible to use another form of ELBO by defining $q(\cdot)$ differently or even other inference techniques such as a MCMC based approaches. While different approaches will incur their corresponding computational effort, it is better to keep the same approach in training GPSSM and computing total mutual information in order to avoid extra computational cost, as we reuse the posterior with trained hyperparameters in estimating the total mutual information.
\end{remark}

%% file: section_experiments.tex
\section{Experiments}\label{section_exp}

In this section, our experiments demonstrate how our proposed AL strategies help in learning GPSSM. From our discussion during previous sections, the two latest work from Buisson et al.\cite{buisson2020actively} and Capone et al.\cite{capone2020localized} assumed known states, so we cannot compare against them in the setting with unobservable states we focus on. Therefore, the reasonable benchmark we can compare to is the usual setting where models are learned with randomly distributed control inputs (e.g. \cite{mattos2016latent}). We first perform an experiment on a simulated function, then on few non-linear physical problems.\footnote{The ground base of GPSSM code is based on Ialongo et al.\cite{ialongo2019overcoming}, re-engineered in \texttt{GPflow} 2.1 \cite{de2017gpflow}.}  The overall goal is to obtain a high accuracy for the GPSSM with as few selected points as possible.

\subsection{Simulated Function}

We first consider a modified kink function based on \cite{ialongo2019overcoming}. Our goal is to learn the dynamics of the system within a range of $\mathcal{X} = \left[-3.0, 1.1\right]$ for the latent states and $\mathcal{C} = \left[0, 1\right]$ for the control. We begin with 5 given points and explore 30 steps. 

\begin{figure}[t!]
\centering
\includegraphics[width=.45\columnwidth]{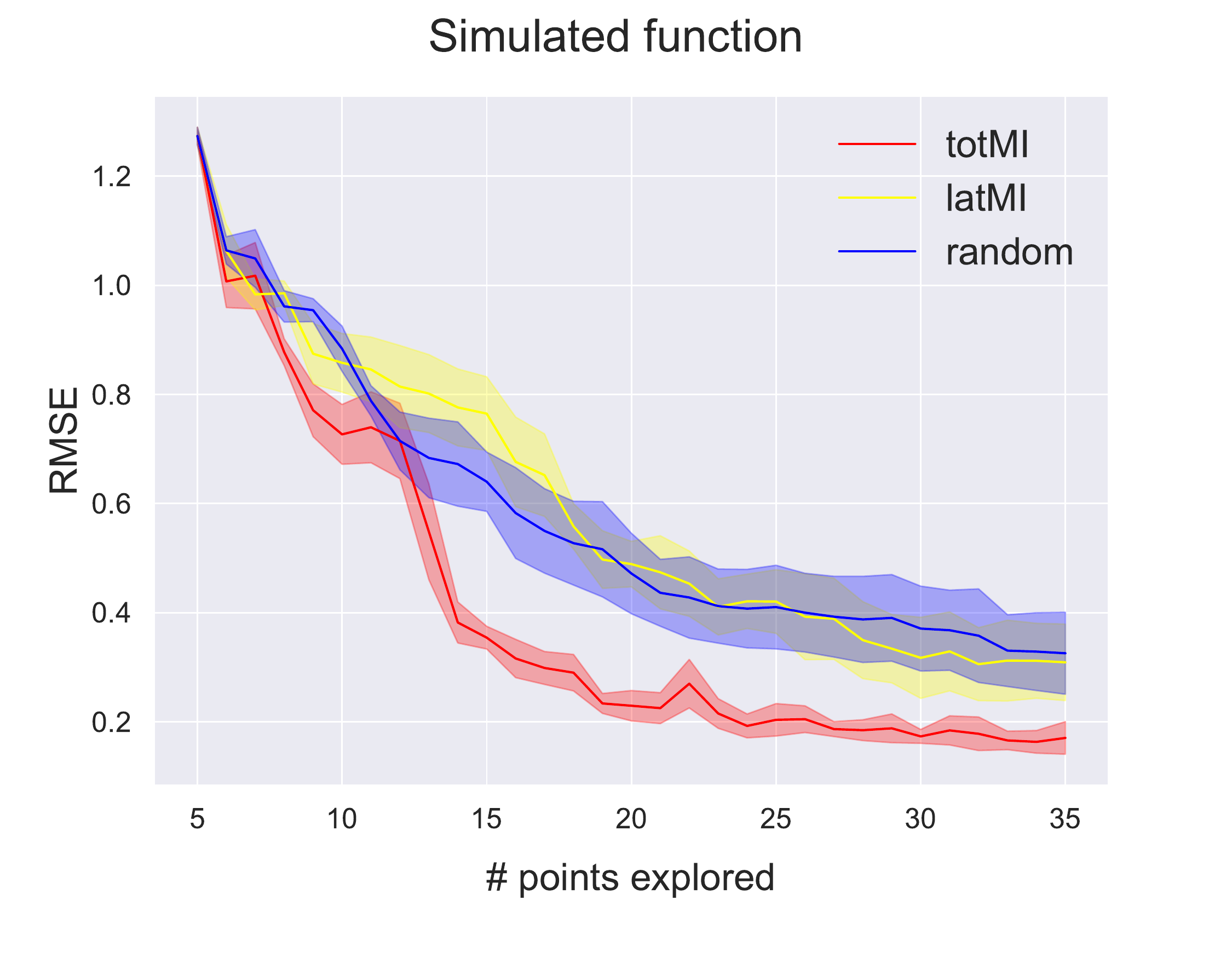}
\includegraphics[width=.45\columnwidth]{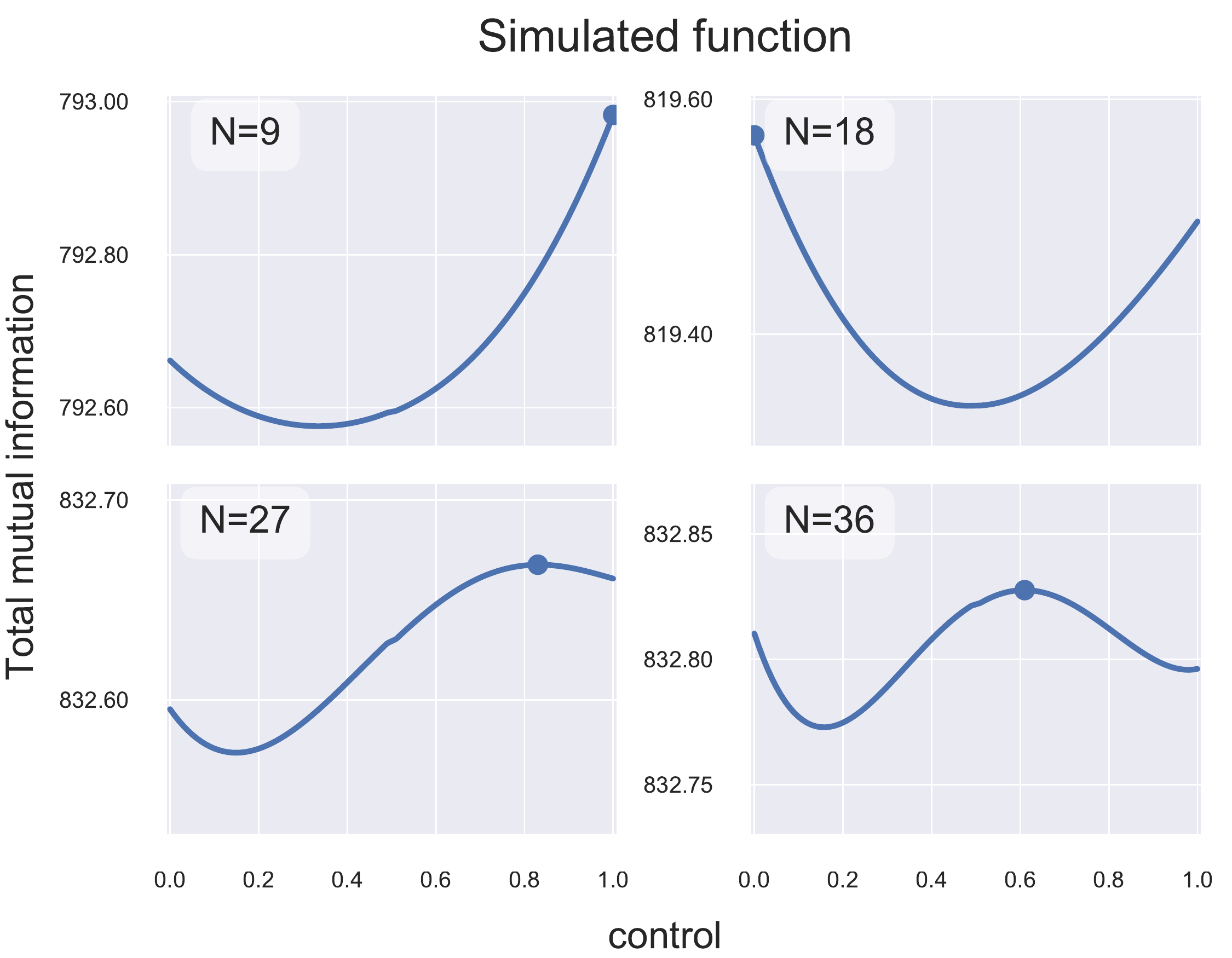}
\caption{The \emph{left} diagram shows the results of the simulated function: Mean and standard deviation of RMSE for 30 exploration steps (starting with 5 initial points) are collected from independent trials. latMI is presented in yellow, totMI in red, and random exploration in blue. The right figures shows where the maximum \emph{total} mutual information is attained when 9, 18, 27, 36 points are explored in one particular trial.}
\label{fig_kink_rmse}
\end{figure}

The left diagram in Figure \ref{fig_kink_rmse} depicts the RMSE of training sessions applying random selection, \emph{latest} mutual information (latMI) and \emph{total} mutual information (totMI) strategy, respectively. Each session consists of independent runs. Soon after collecting the first control input, both AL show results not worse than random but latMI is not significant, while totMI is more profound.

To illustrate why we need our AL strategies, especially totMI, we also present a snapshot of the position of the maximum totMI as we explore in the right of Figure \ref{fig_kink_rmse} in a trial. The selected diagram indicates that there is no trivial most informative control where we can attain maximum totMI. Therefore, optimisation in equation \ref{opt_c_MI_all} (or \ref{opt_c_MI_latest}) is necessary. These diagrams could change as we run different trials, and such evaluation will be more complicated as we increase the dimension of controls or the dynamics become more complex.

\subsection{Pendulum and Cart-Pole}

In these experiments, our aim is to learn the physics of the pendulum and cart-pole by actively controlling the input torque and the force of the cart, respectively. We evaluate the state-space model's accuracy via the angular position obtained from the model against the ground-truth, which is calculated via the equation of motions. The description of the physics of both pendulum and cart-pole, as well as the experimental settings, are presented in supplementary materials. 

Figure \ref{fig_pendulum_rmse} depicts the RMSE of training sessions applying random selection, \emph{latest} mutual information and \emph{total} mutual information strategy for pendulum (left) and cart-pole (right), respectively. Both experiments consist of independent runs. For the pendulum, both totMI and latMI strategy show advantages against the random exploration. However, totMI shows even better results as the exploration proceeds. For cart-pole, all strategies have, in the initial exploration phase, a raise in RMSE, which is subsequently reduced, as the model becomes more stable. latMI is indeed unable to attain noticeable improvement but totMI strategy still shows favourable performance via a consistently lower RMSE.
\begin{figure}[t!]
\centering
\includegraphics[width=.45\columnwidth]{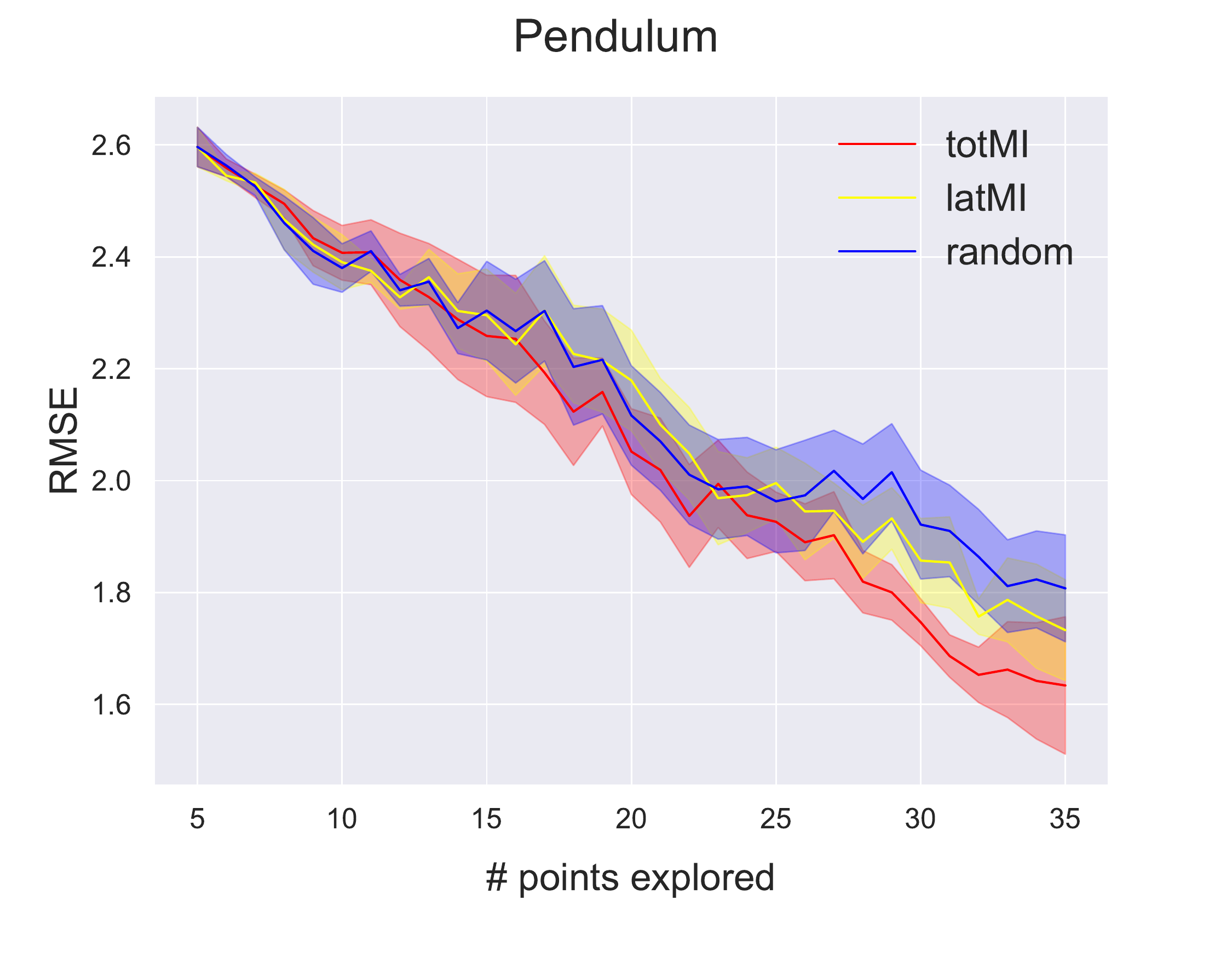}
\includegraphics[width=.45\columnwidth]{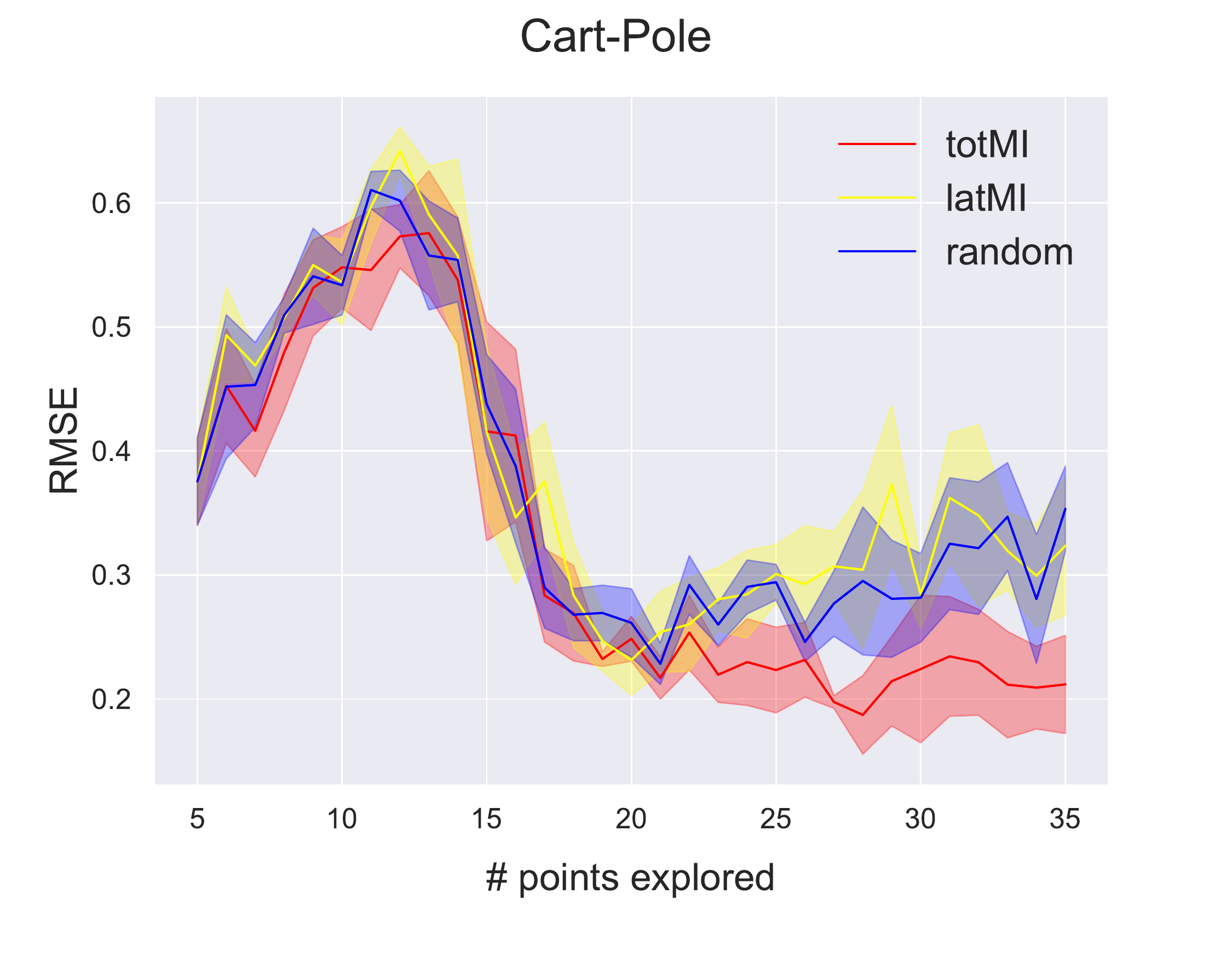}
\caption{The left diagram shows the results of the pendulum: Mean and standard deviation of RMSE for 30 exploration steps (starting with 5 initial points) are collected from independent trials. latMI is presented in yellow, totMI in red, and random exploration in blue. The right diagram shows the results of the cart-pole: Again, mean and standard deviation of RMSE for 30 exploration steps (starting with 5 initial points) are collected from independent trials.}
\label{fig_pendulum_rmse}
\end{figure}


Since the physics of both systems are common enough that we can actually measure the states, we also compare our work directly with the AL strategy by Capone et al.\cite{capone2020localized}, even though the fairness of such a comparison is debatable. We refer to the supplementary materials for a detailed discussion. 




\subsection{Twin-Rotor Aerodynamical System}

\begin{figure}[t!]
\centering
\includegraphics[width=.45\columnwidth]{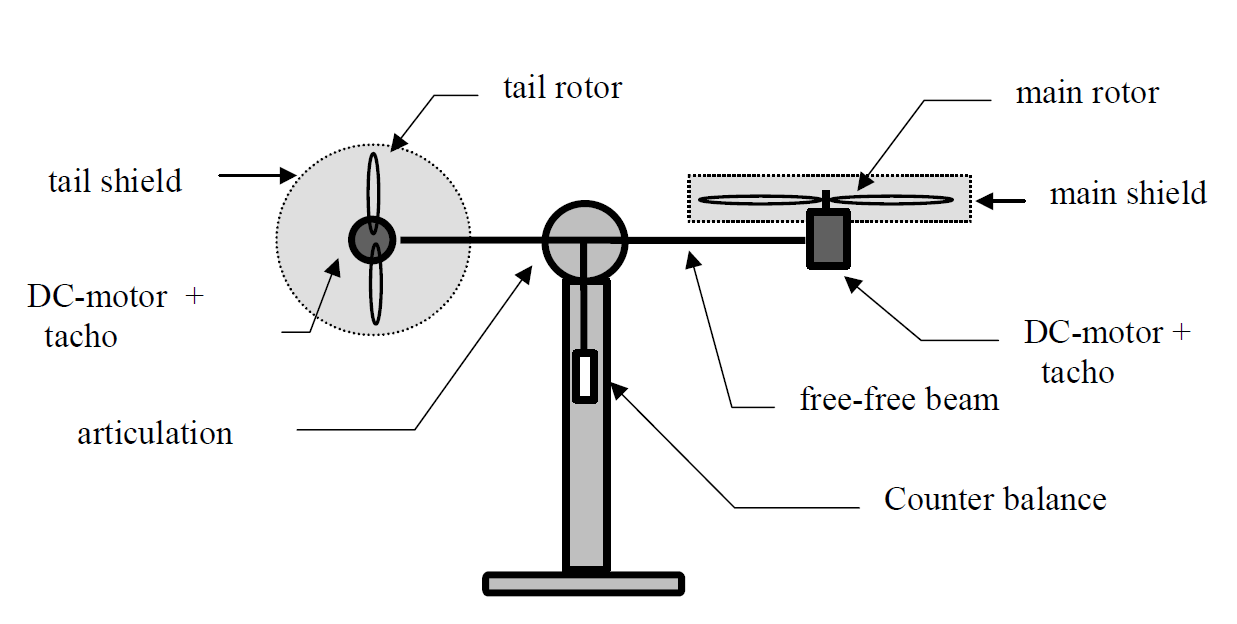}
\includegraphics[width=.45\columnwidth]{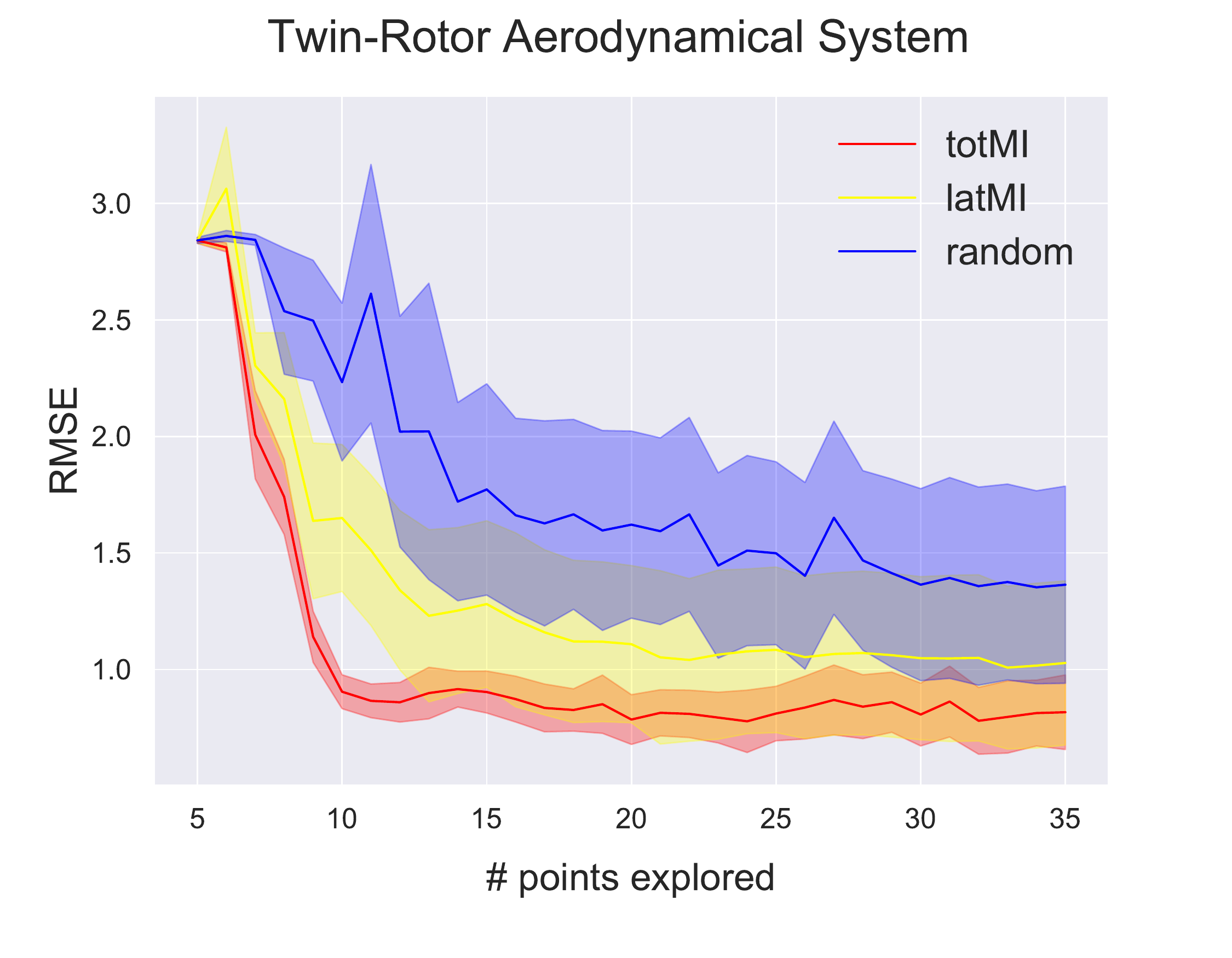}
\caption{The left diagram shows an example of a Twin-Rotor Aerodynamical System, taken from \cite{anony06}. The right diagram shows the results of the Twin-Rotor Aerodynamical System: Mean and standard deviation of RMSE for 30 exploration steps (starting with 5 initial points) are collected from independent trials. latMI is presneted in yellow; totMI in red and random exploration in blue.}
\label{fig_model_tras}
\end{figure}

As a more realistic setting, we test our AL strategy on a Twin-Rotor Aerodynamical System (TRAS). This is a typical design for control experiments, and its behaviour resembles that of a helicopter. There are two rotors -- one horizontal and another vertical -- joined by a beam. There is also a counter-weight, which determines a stable equilibrium position. When the system is switched off, the main rotor is lowered. The motion of the system is controlled by the two motor supply voltages to each motor. We demonstrate a simple example of such a system in the left diagram of Figure \ref{fig_model_tras}.

Our aim is to learn the physics of the TRAS by actively control the two voltages to the system. We evaluate the model's accuracy via the angular position measured from the model against which is calculated via the equation of motions on both rotors. The physics of the TRAS is based on the description by \cite{anony06} and we leave the details in supplementary materials. The right diagram of Figure \ref{fig_model_tras} depicts the RMSE of training sessions applying random selection, latMI and totMI strategy, respectively. Each session consists of independent runs. Both latMI and totMI strategies show advantage against random exploration in presence of more sophisticated dynamics. However, GPSSM with totMI reaches an acceptable accuracy with much lower number of data points.

%% file: section_conclusion.tex
\section{Discussion}\label{section_concl}

We have studied AL in GPSSM by proposing a tractable mutual information estimate, in order to select the most informative control inputs to maximise our model accuracy as we explore. Both \emph{latest} and \emph{total} mutual information are proposed. The empirical results show that \emph{total} mutual information strategy outperforms the other approach and random exploration, which shows that \emph{total} mutual information should be adopted. There is a promising potential of our approach towards more complex industrial settings. Since we leverage ELBO to derive our estimates, with suitable modifications, we also expect our propositions to be applicable to future work on inference of GPSSM and other probabilistic models with latent space. 

Another direction of extension is to look into the use of maximal mutual information in reinforcement learning (RL) community, as such strategy has been used in planning and control. For instance, Ding et al.\cite{ding2020mutual} applied such strategy in Partially Observable Markov Decision Process (POMDP) framework in order to efficiently learn the RL model. However, there are key differences between the two themes. First, the word \emph{control} is interpreted very differently. In GPSSM, this refers to an augmented latent state and each input is simply a deterministic real number whereas in POMDP, this is a mapping between an action taken in certain state to maximise the expected return or reward. Second, AL in GPSSM has a very specific problem formulation whereas efficient learning in POMDP is a general framework with a large room of freedom to specify different components, leading to various research topics (e.g. \cite{engel2005reinforcement, grande2014sample, berkenkamp2017safe}).

%% file: supplementary.tex
\newpage
\section{Supplementary Materials}
%
\subsection{Proof of Propositions}

\textbf{Proposition 1.}
Given the definition of GPSSM from equation 1 - 4, as well as the notation of $\mathcal{M}_t$ and $\mathcal{K}_t$ defined in equation 7. We define the approximation of the following integrals as
\begin{equation*}
\begin{aligned}
\mathcal{N}(\bm{f}_1 \vert M_1, V_1) := \mathcal{N}(\bm{f}_1 \vert M(\bm{\mu}_0, \bm{\Sigma}_0), V(\bm{\mu}_0, \bm{\Sigma}_0)) \approx \int \mathcal{N}(\bm{f}_1 \vert \mathcal{M}_0, \mathcal{K}_0) \mathcal{N}(\bm{x}_0 \vert \bm{\mu}_0, \bm{\Sigma}_0) d\bm{x}_0
\end{aligned}
\end{equation*}
and, recursively, for all $t = T, T+1, \cdots, T+N$,
\begin{equation*}
\mathcal{N}(\bm{f}_t \vert M_t, V_t) \approx \int \mathcal{N}(\bm{f}_t \vert \mathcal{M}_{t-1}, \mathcal{K}_{t-1}) \mathcal{N}(\bm{x}_{t-1} \vert M_{t-1}, V_{t-1} + \bm{Q}) d \bm{x}_{t-1}.
\end{equation*}

Then, the latest mutual information is approximately
\begin{equation}\label{eq_prop_mi_latest_supp}
I(\bm{y}_t; \bm{f}_t) \approx \frac{1}{2} \log\left(\frac{\det(\bm{R} + \bm{C}(V_t + \bm{Q})\bm{C}^T)}{\det(\bm{R} + \bm{CQC}^T)}\right).
\end{equation}

Before the proof, we first elaborate the Gaussian approximation technique from Girard\cite{girard2004approximate}. 

For two random variables $\bm{x}$ and $\bm{y}$ with $p(\bm{y} \vert \bm{x}) = \mathcal{N}(\bm{y} \vert \mu(\bm{x}), \sigma^2(\bm{x}))$ and $p(\bm{x} \vert \bm{u}, \bm{\Sigma}_{\bm{x}}) = \mathcal{N}(\bm{x} \vert \bm{u}, \bm{\Sigma}_{\bm{x}})$ for some mean $\bm{u}$ and variance $\Sigma_{\bm{x}}$, the Gaussian approximation technique from \cite{girard2004approximate} yields to
\begin{equation*}
\int p(\bm{y} \vert \bm{x}) p(\bm{x} \vert \bm{u}, \bm{\Sigma}_{\bm{x}}) d\bm{x} \approx \mathcal{N}(M(\bm{u}, \bm{\Sigma}_{\bm{x}}), V(\bm{u}, \bm{\Sigma}_{\bm{x}})),
\end{equation*}
where 
\begin{equation*}
\begin{aligned}
M(\bm{u}, \bm{\Sigma}_{\bm{x}}) &= \int \mu(\bm{x}) p(\bm{x} \vert \bm{u}, \bm{\Sigma}_{\bm{x}}) d\bm{x}, \\
V(\bm{u}, \bm{\Sigma}_{\bm{x}}) &= \int \sigma^2(\bm{x}) p(\bm{x} \vert \bm{u}, \bm{\Sigma}_{\bm{x}}) d\bm{x} + \int \mu^2(\bm{x}) p(\bm{x} \vert \bm{u}, \bm{\Sigma}_{\bm{x}}) d\bm{x} - M(\bm{u}, \bm{\Sigma}_{\bm{x}})^2.
\end{aligned}
\end{equation*}

In particular, if $\mu(\bm{x})$ is linear (e.g. $\bm{\mu}(\bm{x}) = \bm{Fx} + \bm{a}$) and $\sigma^2(\bm{x})$ can be represented as a matrix $\bm{A}$. The integral reduces to the exact formula of Gaussian integral
\begin{equation*}
\int \mathcal{N}(\bm{y} \vert \bm{Fx} + \bm{a}, \bm{A}) \mathcal{N}(\bm{x} \vert \bm{u}, \bm{\Sigma}_{\bm{x}}) d\bm{x} = \mathcal{N}(\bm{y} \vert \bm{a} + \bm{Fu}, \bm{A} + \bm{F}\bm{\Sigma}_{\bm{x}}\bm{F}^T).
\end{equation*}

In the context of GP, $M(\cdot, \cdot)$ and $V(\cdot, \cdot)$ can be written based on the expectation of the kernel, which is 
\begin{equation*}
\begin{aligned}
M(\cdot, \cdot) &= \sum_{i=1}^{t} \beta_i E_{\bm{x}}\left[k(\bm{x}, \bm{x}_i)\right], \\
V(\cdot, \cdot) &= E_{\bm{x}}\left[k(\bm{x}, \bm{x})\right] - \sum_{i, j = 1}^{t} (K^{-1}_{ij} - \beta_i \beta_j) E_{\bm{x}}\left[k(\bm{x}, \bm{x}_i) k(\bm{x}, \bm{x}_j)\right] - M(\bm{u}, \bm{\Sigma}_{\bm{x}})^2.
\end{aligned}
\end{equation*}
where $\bm{\beta} = K^{-1} \bm{y}$ and $\bm{\beta} = \lbrace \beta_i \rbrace^t_{i=1}$ and $K$ refers to the matrix of all covariances of the GP. There are three terms we need to evaluate, denoted by
\begin{equation*}
\begin{aligned}
l &:= E_{\bm{x}}\left[k(\bm{x}, \bm{x})\right], \\
l_i &:= E_{\bm{x}}\left[k(\bm{x}, \bm{x}_i)\right], \\
l_{ij} &:= E_{\bm{x}}\left[k(\bm{x}, \bm{x}_i) k(\bm{x}, \bm{x}_j)\right].
\end{aligned}
\end{equation*}

This can be further simplified when the covariance function has a specific form. For example, if the kernel is squared exponential,
\begin{equation*}
k(\bm{x}_i, \bm{x}_j) = \sigma^2 \exp\left[-\frac{1}{2} (\bm{x}_i - \bm{x}_j)^T \Lambda^{-1} (\bm{x}_i - \bm{x}_j), \right]
\end{equation*}
which can be rewritten as 
\begin{equation*}
k(\bm{x}_i, \bm{x}_j) = c \mathcal{N}\left( \bm{x}_i \vert \bm{x}_j, \Lambda \right), 
\end{equation*}
with $c = (2\pi)^{d_x / 2} \vert \Lambda \vert^{1/2} \sigma^2$, we can show that
\begin{equation*}
\begin{aligned}
l &= l = \sigma^2, \\
l_i &= c \mathcal{N}\left( \bm{u} \vert \bm{x}_i, \Lambda + \bm{\Sigma}_{\bm{x}} \right), \\
l_{ij} &:= c \mathcal{N}(\bm{x}_i \vert \bm{x}_j, 2 \Lambda) \mathcal{N}(\bm{u} \vert \frac{1}{2}(\bm{x}_i + \bm{x}_j), \frac{1}{2}\Lambda + \bm{\Sigma}_{\bm{x}}).
\end{aligned}
\end{equation*}

\begin{proof}
We begin by using the relationship $I(\bm{y}_t ; \bm{f}_t) = h(\bm{y}_t) - h(\bm{y}_t \vert \bm{f}_t)$. Note that ELBO $\mathcal{L}_t$, as presented in equation 9, gives a bound of $p(\bm{y}_{1:t})$ but not a specific $p(\bm{y}_t)$. There is no clear and simple relationship between one observation and the joint of all observations, so we require an approach which allows us to estimate $p(\bm{y}_t)$ directly. While there could be different estimates available, we adopt the approximate Gaussian integral from \cite{girard2004approximate} because of reason (1) and (2) stated in section 4. 

First, we express $p(\bm{y}_t)$ as
\begin{equation*}
\begin{aligned}
p(\bm{y}_t) &= \int p(\bm{y}_t, \bm{x}_{0:t}, \bm{f}_{1:t}) d\bm{x}_{0:t} d\bm{f}_{1:t} \\
&= \int p(\bm{y}_t \vert \bm{x}_{0:t}, \bm{f}_{1:t}) p(\bm{x}_t \vert \bm{x}_{0:t-1}, \bm{f}_{1:t}) p(\bm{f}_t \vert \bm{x}_{0:t-1}, \bm{f}_{1:t-1}) p(\bm{x}_{0:t-1}, \bm{f}_{1:t-1}) d\bm{x}_{0:t} d\bm{f}_{1:t} \\
&= \int p(\bm{y}_t \vert \bm{x}_t) p(\bm{x}_t \vert \bm{f}_t) p(\bm{f}_t \vert \bm{x}_{0:t-1}, \bm{f}_{1:t-1}) p(\bm{x}_{0:t-1}, \bm{f}_{1:t-1}) d\bm{x}_{0:t} d\bm{f}_{1:t} \\
&= \int p(\bm{y}_t \vert \bm{x}_t) \prod_{i=1}^{t} p(\bm{x}_i \vert \bm{f}_i) p(\bm{f}_i \vert \bm{x}_{0:i-1}, \bm{f}_{1:i-1}) p(\bm{x}_0) d\bm{x}_{0:t} d\bm{f}_{1:t} \\
&= \int \mathcal{N}(\bm{y}_t \vert \bm{Cx}_t + \bm{d}, \bm{R}) \prod_{i=1}^{t} \mathcal{N}(\bm{x}_i \vert \bm{f}_i, \bm{Q}) \mathcal{N}(\bm{f}_i \vert \mathcal{M}_{i-1}, \mathcal{K}_{i-1}) \mathcal{N}(\bm{x}_0 \vert \bm{\mu}_0, \bm{\Sigma}_0) d\bm{x}_{0:t} d\bm{f}_{1:t},
\end{aligned}
\end{equation*}
where the third line and terms in the last line follow by the definition of GPSSM in the main text (equation 1-4 and 7). Note that in the forth line, $\bm{f}_{1:0} \equiv \emptyset$.

The next step is to integrate the expression in the order of $\bm{x}_0 \rightarrow \bm{f}_1 \rightarrow \bm{x}_1 \cdots \rightarrow \bm{f}_t \rightarrow \bm{x}_t$. Starting with
\begin{equation*}
\int \mathcal{N}(\bm{f}_1 \vert \mathcal{M}_0, \mathcal{K}_0) \mathcal{N}(\bm{x}_0 \vert \bm{\mu}_0, \bm{\Sigma}_0) d\bm{x}_0
\end{equation*}
and since $\mathcal{M}_0$, which depends on $\bm{x}_0$, is in general non-linear, we cannot use the identity of Gaussian integral and this is where we apply the Gaussian approximation by \cite{girard2004approximate}. By interlacing the exact and approximate Gaussian integral, we first have
\begin{equation*}
\int \mathcal{N}(\bm{f}_1 \vert \mathcal{M}_0, \mathcal{K}_0) \mathcal{N}(\bm{x}_0 \vert \bm{\mu}_0, \bm{\Sigma}_0) d\bm{x}_0 \approx \mathcal{N}(\bm{f}_1 \vert M(\bm{\mu}_0, \bm{\Sigma}_0), V(\bm{\mu}_0, \bm{\Sigma}_0)) =: \mathcal{N}(\bm{f}_1 \vert M_1, V_1).
\end{equation*}

Then, the next integration is an exact Gaussian integral
\begin{equation*}
\int \mathcal{N}(\bm{x}_1 \vert \bm{f}_1, \bm{Q}) \mathcal{N}(\bm{f}_1 \vert M_1, V_1) d\bm{f}_1 = \mathcal{N}(\bm{x}_1 \vert M_1, \bm{Q} + V_1).
\end{equation*}

The process continues until the term $M_t$ and $V_t$ are defined (recall that for $\mathcal{M}_t$ and $\mathcal{K}_t$, these two terms depends on $\bm{x}_{0:t}$ and $\bm{f}_{1:t}$):
\begin{equation*}
\begin{aligned}
\int \mathcal{N}(\bm{f}_2 \vert \mathcal{M}_1, \mathcal{K}_1) \mathcal{N}(\bm{x}_1 \vert M_1, \bm{Q} + V_1) d\bm{x}_1 &\approx \mathcal{N}(\bm{f}_2 \vert M(M_1, \bm{Q} + V_1), V(M_1, \bm{Q} + V_1)) \\
&=: \mathcal{N}(\bm{f}_2 \vert M_2, V_2), \\
\int \mathcal{N}(\bm{x}_2 \vert \bm{f}_2, \bm{Q}) \mathcal{N}(\bm{f}_2 \vert M_2, V_2) d\bm{f}_2 &= \mathcal{N}(\bm{x}_2 \vert M_2, \bm{Q} + V_2), \\
&\vdots \\
\int \mathcal{N}(\bm{f}_t \vert \mathcal{M}_{t-1}, \mathcal{K}_{t-1}) \mathcal{N}(\bm{x}_{t-1} \vert M_{t-1}, \bm{Q} + V_{t-1}) d\bm{x}_{t-1} &\approx \mathcal{N}(\bm{f}_t \vert M(M_{t-1}, \bm{Q} + V_{t-1}), V(M_{t-1}, \bm{Q} + V_{t-1})) \\
&= \mathcal{N}(\bm{f}_t \vert M_t, V_t) \\
\int \mathcal{N}(\bm{x}_t \vert \bm{f}_t, \bm{Q}) \mathcal{N}(\bm{f}_t \vert M_t, V_t) d\bm{f}_t &= \mathcal{N}(\bm{x}_t \vert M_t, \bm{Q} + V_t).
\end{aligned}
\end{equation*}

The last integration with respect to $\bm{x}_t$, which is an exact Gaussian Integral, is given by
\begin{equation*}
\int \mathcal{N}(\bm{y}_t \vert \bm{Cx}_t + \bm{d}, \bm{R}) \mathcal{N}(\bm{x}_t \vert M_t, \bm{Q} + V_t) d\bm{x}_t = \mathcal{N}(\bm{y}_t \vert \bm{C} M_t + \bm{d}, \bm{R} + \bm{C}(\bm{Q} + V_t)\bm{C}^T).
\end{equation*}

Hence, it shows that approximately, $p(\bm{y}_t) \approx \mathcal{N}(\bm{y}_t \vert \bm{C} M_t + \bm{d}, \bm{R} + \bm{C}(\bm{Q} + V_t)\bm{C}^T)$. Since this is a normally distributed expression, the entropy of which is given by
\begin{equation*}
h(\bm{y}_t) \approx \frac{1}{2} \logdet\left(2\pi e (\bm{R} + \bm{C}(\bm{Q} + V_t)\bm{C}^T)\right)
\end{equation*}

It remains to compute $h(\bm{y}_t \vert \bm{f}_t)$. This term can be directly computed, as
\begin{equation*}
\begin{aligned}
p(\bm{y}_t \vert \bm{f}_t) &= \int p(\bm{y}_t, \bm{x}_t \vert \bm{f}_t) d\bm{x}_t \\
&= \int p(\bm{y}_t \vert \bm{x}_t, \bm{f}_t) p(\bm{x}_t \vert \bm{f}_t) d\bm{x}_t \\
&= \int p(\bm{y}_t \vert \bm{x}_t) p(\bm{x}_t \vert \bm{f}_t) d\bm{x}_t \\
&= \int \mathcal{N}\left(\bm{y}_t \vert \bm{Cx}_t + \bm{d}, \bm{R} \right) \mathcal{N}\left(\bm{x}_t \vert \bm{f}_t, \bm{Q} \right) d\bm{x}_t \\
&= \mathcal{N} \left(\bm{y}_t \vert \bm{Cf}_t + \bm{d}, \bm{R} + \bm{CQC}^T \right).
\end{aligned}
\end{equation*}

So, the entropy of this distribution is given by
\begin{equation*}
h(\bm{y}_t \vert \bm{f}_t) = \frac{1}{2} \logdet\left(2\pi e (\bm{R} + \bm{CQC}^T)\right).
\end{equation*}

Finally, we obtain the approximation of the mutual information between the latest observation and prediction as stated in the proposition, given by
\begin{equation*}
\begin{aligned}
I(\bm{y}_t ; \bm{f}_t) &\approx \frac{1}{2} \logdet\left(2\pi e (\bm{R} + \bm{C}(\bm{Q} + V_t)\bm{C}^T)\right) - \frac{1}{2} \logdet\left(2\pi e (\bm{R} + \bm{CQC}^T)\right) \\
&= \frac{1}{2}\log\left(\frac{\det(\bm{R} + \bm{C}(\bm{Q} + V_t)\bm{C}^T)}{\det(\bm{R} + \bm{CQC}^T)}\right).
\end{aligned}
\end{equation*}
\end{proof}

\medskip

\begin{flushleft}
\textbf{Proposition 2.} 
Given the definition of GPSSM from equation 1 - 4, as well as the expression of the ELBO $\mathcal{L}_t$ from equation 11, for $t = T, T+1, \cdots, T+N$, if $S$ samples are drawn, the $s$-th sample estimate ($s = 1, \cdots, S$) of the mutual information between observations $\bm{y}_{1:t}$ and the prediction $\bm{f}_{1:t}$, denoted by $i_s$, is bounded by
\begin{equation}\label{eq_prop_totalMI_persample_supp}
i_s \leq  \sum^t_{i=1} \log(\mathcal{N}_s(\bm{y}_i \vert \bm{Cf}_i + \bm{d}, \bm{R} + \bm{CQC}^T)) - \mathcal{L}_{t, s},
\end{equation}
where $\mathcal{N}_s(\cdot)$ and $\mathcal{L}_{t, s}$ are the $s$-th sample estimate of the normal distribution and ELBO $\mathcal{L}_t$, respectively. The total mutual information is then approximated by $I(\bm{y}_{1:t}; \bm{f}_{1:t}) \approx \frac{1}{S}\sum^S_{s=1} i_s$.
\end{flushleft}

\begin{proof}
Again, we first use $I(\bm{y}_{1:t}; \bm{f}_{1:t}) = h(\bm{y}_{1:t}) - h(\bm{y}_{1:t} \vert \bm{f}_{1:t})$. Since we have the estimate for the probability density function (pdf), a generic Monte-Carlo estimate of the entropy (e.g. \cite{lombardi2016nonparametric}) is $h(\cdot) = \frac{1}{S} \sum^S_{i=1} \log\left(\frac{1}{\hat{p}_s(\cdot)}\right)$, where $S$ is the number of samples drawn and $\hat{p}_s(\cdot)$ is the estimate of the pdf of the random variables. With two estimates of pdf formulas, the total mutual information can be estimated by
\begin{equation*}
\begin{aligned}
I(\bm{y}_{1:t}; \bm{f}_{1:t}) &\approx \frac{1}{S} \sum^S_{s=1} \log\left(\frac{1}{\hat{p}_s(\bm{y}_{1:t})}\right) - \frac{1}{S} \sum^S_{s=1} \log\left(\frac{1}{\hat{p}_s(\bm{y}_{1:t} \vert \bm{f}_{1:t})}\right) \\
&\approx \frac{1}{S} \sum^S_{s=1} \left( \log\left(\hat{p}_s(\bm{y}_{1:t} \vert \bm{f}_{1:t}) \right) - \log\left(\hat{p}_s(\bm{y}_{1:t}) \right) \right).
\end{aligned}
\end{equation*}

This shows that for the $s$-th sample, $s = 1, \cdots, S$, the total mutual information estimate can be defined and bounded by
\begin{equation*}
\begin{aligned}
i_s &:= \log\left(\hat{p}_s(\bm{y}_{1:t} \vert \bm{f}_{1:t}) \right) - \log\left(\hat{p}_s(\bm{y}_{1:t}) \right) \\
&\leq \log\left(\hat{p}_s(\bm{y}_{1:t} \vert \bm{f}_{1:t}) \right) - \mathcal{L}_{t, s},
\end{aligned}
\end{equation*}
where the upper bound is based on the inference scheme we have in training the GPSSM, as we have derived an ELBO $\mathcal{L}_t$ (equation 9) and we use the fact that $\log\left(p(\bm{y}_{1:t})\right) \geq \mathcal{L}_t$.

The remaining part is to compute $\hat{p}_s(\bm{y}_{1:t} \vert \bm{f}_{1:t})$ and this term can be exactly derived based on the definition of the GPSSM. The outcome is a product of normal random variables given by
\begin{equation*}
\begin{aligned}
\hat{p}_s(\bm{y}_{1:t} \vert \bm{f}_{1:t}) &= \int \hat{p}_s(\bm{y}_{1:t}, \bm{x}_{0:t} \vert \bm{f}_{1:t}) d\bm{x}_{0:t} \\
&= \int \hat{p}_s(\bm{y}_{t} \vert \bm{y}_{1:t-1}, \bm{x}_{0:t}, \bm{f}_{1:t}) \hat{p}_s(\bm{x}_{t} \vert \bm{y}_{1:t-1}, \bm{x}_{0:t-1}, \bm{f}_{1:t}) \hat{p}_s(\bm{y}_{1:t-1}, \bm{x}_{0:t-1} \vert \bm{f}_{1:t}) d\bm{x}_{0:t} \\
&= \int \prod_{i=1}^{t} \hat{p}_s(\bm{y}_i \vert \bm{x}_i) \hat{p}_s(\bm{x}_i \vert \bm{f}_i) d\bm{x}_{1:t},
\end{aligned}
\end{equation*}
where the last line is based on using the definition of GPSSM recursively. The remaining derivation is purely mathematical techniques involving Fubini's theorem and Gaussian integral.
\begin{equation*}
\begin{aligned}
\int \prod_{i=1}^{t} \hat{p}_s(\bm{y}_i \vert \bm{x}_i) \hat{p}_s(\bm{x}_i \vert \bm{f}_i) d\bm{x}_{1:t} &= \int \prod_{i=1}^{t} \mathcal{N}_s(\bm{y}_i \vert \bm{Cx}_i + \bm{d}, \bm{R}) \mathcal{N}_s(\bm{x}_i \vert \bm{f}_i, \bm{Q}) d\bm{x}_{1:t} \\
&= \prod_{i=1}^{t} \left( \int  \mathcal{N}_s(\bm{y}_i \vert \bm{Cx}_i + \bm{d}, \bm{R}) \mathcal{N}(\bm{x}_i \vert \bm{f}_i, \bm{Q}) d\bm{x}_i \right) \\
&= \prod_{i=1}^{t} \mathcal{N}_s(\bm{y}_i \vert \bm{Cf}_i + \bm{d}, \bm{R} + \bm{CQC}^T).
\end{aligned}
\end{equation*}
With products becomes summations by taking logarithm, we arrive proposition 4.1.
\end{proof}

\begin{remark}
If the dimension of the observation is 1 (i.e. $d_y = 1$), the product of normal random variables can be further simplified. This can be written in a closed form based on Meijer G-function \cite{springer1970distribution}. Recent work showed that it is possible to compute the cumulative distribution function for such distribution as well (e.g. \cite{stojanac2017products}). We postulate that there will be development on the entropy for such distribution from the mathematics community which could further simplify our expression.
\end{remark}

\subsection{Experimental Details}
Here we supplement the experimental section in the main text for further details and results. We also present the equations of motion for the physical problems studied for completeness.

\subsubsection{Simulated function}
The transition function of the modified kink function, based on Ialongo et al.\cite{ialongo2019overcoming}, is given by
\begin{equation}
y_{t} = c_t + (y_{t - 1} + 0.2) \left(1 - \frac{5}{1 + \exp(-2 y_{t-1})} \right),
\end{equation}
where $y_t$ and $c_t$ are the target observation and control input, respectively. We set the latent state $x_t = y_t$ for simplicity.

\subsubsection{Pendulum and Cart-Pole - the physics}
The observed variable here is the angle of the pendulum $\theta$ (i.e. $y_t \equiv \theta_t$) and the control $c$ is the torque acting on the pendulum. The equations of motion are given by
\begin{equation}
\begin{aligned}
\frac{d \theta}{d t} &= \dot{\theta}, \\
\frac{d \dot{\theta}}{d t} &= \ddot{\theta} = \frac{1}{I} \left(-b \dot{\theta}  - mgl \sin(\theta) + c \right).
\end{aligned}
\end{equation}
The value of the parameters are $g = 9.81$ ms$^{-2}$, $m = 0.1$ kg, $l = 1$ m, $b = 0.05$ and $I = ml^2$.

The cart-pole we used is based on Florian\cite{florian2007correct}. The observed variables are the angle of the pendulum $\theta$ and the position of the cart $x$. The control $c$ is the force acting on the cart. The equations of motion are given by 
\begin{equation}
\begin{aligned}
N_c &= (m_c + m_p) g - m_p l \left( \ddot{\theta} \sin(\theta) + \dot{\theta}^2 \cos(\theta) \right), \\
\frac{d \theta}{d t} &= \dot{\theta}, \\
\frac{d \dot{\theta}}{d t} &= \ddot{\theta} = \frac{g \sin(\theta) + \cos(\theta) \left(\frac{-c - m_p l \dot{\theta}^2 \left( \sin(\theta) + \mu_c \sgn(N_c \dot{x}) \cos(\theta) \right)}{m_c + m_p} + \mu_c g \sgn(N_c \dot{x}) \right) - \frac{\mu_p \dot{\theta}}{m_p l}}{l \left( \frac{4}{3} - \frac{m_p \cos(\theta)}{m_c + m_p} \left( \cos(\theta) - \mu_c \sgn(N_c \dot{x}) \right) \right)}, \\
\frac{d x}{d t} &= \dot{x}, \\
\frac{d \dot{x}}{d t} &= \ddot{x} = \frac{c + m_p l \left( \dot{\theta}^2 \sin(\theta) - \ddot{\theta} \cos(\theta) \right) - \mu_c N_c \sgn(N_c \dot{x})}{m_c + m_p}.
\end{aligned}
\end{equation}
The value of the parameters are $g = 9.81$ ms$^{-2}$, $m_p = 0.1$ kg, $m_c = 0.5$ kg, $l = 0.5$ m, $\mu_c = 0.05$ and $\mu_p = 0.01$.

\subsubsection{Twin-Rotor Aerodynamical System (TRAS) - the physics}
Based on Petrov et al.\cite{petkov2008robust}, this system consists of two rotors, one horizontal and one vertical, which resembles a simple helicopter. Therefore, the observed variables are azimuth (horizontal) angle $\alpha_h$ and pitch (vertical) angle $\alpha_v$. The equations of motion are given by
\begin{equation}
\begin{aligned}
\frac{d \alpha_v}{d t} &= \dot{\alpha_v}, \\
\frac{d \dot{\alpha_v}}{d t} &= \ddot{\alpha_v} = \frac{M_v}{J_v}, \\
\frac{d \alpha_h}{d t} &= \dot{\alpha_h}, \\
\frac{d \dot{\alpha_h}}{d t} &= \ddot{\alpha_h} = \frac{M_h}{J_h},
\end{aligned}
\end{equation}
where $M_v$ and $M_h$ are the moment of forces in the vertical and horizontal plane, respectively. Here, we use subscript $v$ for the vertical plane and $h$ for the horizontal plane, unless otherwise specified. $J_v$ and $J_h$ are the moments of inertia relative to the two axis. The controls are the two torques acting on the two rotors, which come down to the voltage applied, denoted by $c_v$ and $c_h$.

The four terms ($M_v$, $M_h$, $J_v$ and $J_h$) can be analysed individually. We reproduce part of section 3 of \cite{anony06} so that this subsection is self-contained to understand the physics of the system and the parameters used. The list of all parameters used are summarised in Table \ref{tras_parameters}.
\begin{table}[t!]
\caption{Twin-Rotor Aerodynamical System Parameters} \label{tras_parameters}
\begin{center}
\begin{tabular}{llll}
\textbf{Symbol}  & \textbf{Value} & \textbf{Units} & \textbf{Description}\\
\hline
$g$ & 9.81 & ms$^{-2}$ & the gravitational acceleration \\
$m_m$ & 0.029&kg & the mass of the main part of the beam \\
$m_{mr}$ & 0.199 &kg & the mass of the main rotor \\
$m_{ms}$ &0.083 &kg & the mass of the shield of the main rotor \\
$m_t$ &0.031 &kg & the mass of the tail part of the beam \\
$m_{tr}$ & 0.154 &kg & the mass of the tail rotor\\
$m_{ts}$ & 0.061 &kg & the mass of the shield of the tail rotor \\
$m_b$ & 0.011 &kg  & the mass of the counter-weight beam \\
$m_{cb}$ & 0.024 &kg & the mass of the counter-weight\\
$l_m$ & 0.202 & m & the length of the main part of the beam\\
$l_t$ & 0.216 & m & the length of the tail part of the beam\\
$l_b$ & 0.15 &m & the length of the counter-weight beam \\
$l_{cb}$ & 0.15 & m & the distance between the counter-weight and the joint\\
$r_{ms}$ & 0.145 &m & the radius of the main shield\\
$r_{ts}$ & 0.1 &m & the radius of the tail shield\\
$k_{fv}$ & 0.013 & Nms $\cdot$ rad$^{-1}$ & the friction coefficient in the vertical axis \\
$k_{fh}$ & 0.006 & Nms $\cdot$ rad$^{-1}$& the friction coefficient in the horizontal axis  \\
$k_{hv}$ & 0.004& Nm & the coefficient of the cross moment from tail rotor to pitch \\
$k_{vh}$ & $-0.018$ & Nm &coefficient of the cross moment from main rotor to azimuth \\
$a_1$ & 0.001 & & constant \\
$a_2$ & 0.01 & & constant \\
\end{tabular}
\end{center}
\end{table}

\begin{itemize}
	\item $M_v$: There are 6 components. That is, $M_v = \sum^6_{i=1} M_{vi}$. Here
	\begin{enumerate}
	\item $M_{v1}$ is the return torque corresponding to the forces of gravity. If we refer to Figure 6 in the main text, there are 8 components of masses and 4 lengths to be considered, which are denoted by the letter $m$ and $l$, respectively (see Table~\ref{tras_parameters} for detailed description).
	
	When we group them together, the equation of motion for this part is 
	\begin{equation*}
	M_{v1} = g \lbrace \left[\left( \frac{m_t}{2} + m_{tr} + m_{ts} \right) l_t + \left( \frac{m_m}{2} + m_{mr} + m_{ms} \right) l_m \right] \cos(\alpha_v) - \left[ \frac{m_b}{2}l_b + m_{cb}l_{cb} \right] \sin(\alpha_v) \rbrace.
	\end{equation*}
	
	\item $M_{v2}$ is the moment of the propulsive force produced by the main motor. Here
	\begin{equation*}
	M_{v2} = l_m F_v(\omega_v),
	\end{equation*}
	where $l_m$ is the length of the main part of the beam. $F_v(\omega_v)$ is the dependence of the propulsive force on the angular velocity of the rotor $\omega_v$. This is a quantity measured empirically but for our simulation purposes, we model this function as follow:
	\begin{equation*}
	\begin{aligned}
	F_v(\omega_v) &= -1.8 \cdot 10^{-18} \omega^5_v - 7.8 \cdot 10^{-16} \omega^4_v + 4.1 \cdot 10^{-11} \omega^3_v + 2.7 \cdot 10^{-8} \omega^2_v + 3.5 \cdot 10^{-5} \omega_v - 0.014, \\
	\omega_v(c_v) &= -5.2 \cdot 10^3 c^7_v - 1.1 \cdot 10^2 c^6_v + 1.1 \cdot 10^4 c^5_v - 1.3 \cdot 10^2 c^4_v - 9.2 \cdot 10^3 c^3_v - 31 c^2_v + 6.1 \cdot 10^3 c_v - 4.5.
	\end{aligned}
	\end{equation*}
	Here we can see the the angular velocity is determined by the control of the voltage.
	
	\item $M_{v3}$ is the moment of the centrifugal forces to the motion of the beam vertically, with further derivations this is given by 
	\begin{equation*}
	M_{v3} = -\dot{\alpha_h}^2 \left[ \left( \frac{m_m}{2} + m_{mr} + m_{ms} \right) + \left( \frac{m_t}{2} + m_{tr} + m_{ts} \right) + m_{cb}l_{cb} + \frac{m_b}{2}l_b \right] \sin(\alpha_v)\cos(\alpha_v)
	\end{equation*}
	
	\item $M_{v4}$ is the moment of friction depending on the angular velocity of the beam horizontally, given by 
	\begin{equation*}
	M_{v4} = -\dot{\alpha_v} f_v,
	\end{equation*}
	where $f_v$ is a constant.
	
	\item $M_{v5}$ is the cross moment from horizontal control input, given by
	\begin{equation*}
	M_{v5} = -c_h k_{hv},
	\end{equation*}
	where $k_{hv}$ is a constant.
	
	\item $M_{v6}$ is the damping torque from the propeller, given by
	\begin{equation*}
	M_{v6} = -a_1 \dot{\alpha_v} \abs(\omega_v),
	\end{equation*} 
	where $a_1$ is a constant.
	\end{enumerate}
	
	\item $J_v$: this term turns out to be independent of the position of the beam. By defining $r_{ms}$ as the radius of the main shield of the rotor, $r_{ts}$ as the radius of the tail shield of the rotor, and using the concept of moment of intertia of sphere and cylinders. The sum of moment of inertia is given by
	\begin{equation*}
	J_v = m_{mr} l^2_m + m_m \frac{l^2_m}{3} + m_{cb} l^2_{cb} + m_b \frac{l^2_b}{3} + m_{tr} l^2_t + m_t \frac{l^2_t}{3} + \frac{m_{ms}}{2}r^2_{ms} + m_{ms} l^2_m + m_{ts}r^2_{ts} + m_{ts} l^2_t.
	\end{equation*}
	
	\item $M_h$: There are 4 components. That is, $M_h = \sum^4_{i=1} M_{hi}$. Here
	\begin{enumerate}
	\item $M_{h1}$ is the moment of force horizontally, given by
	\begin{equation*}
	M_{h1} = l_t F_h(\omega_h) \cos(\alpha_v),
	\end{equation*}
	where $F_h(\omega_h)$ is the dependence of the propulsive force on the angular velocity of the rotor $\omega_h$. Again, this is a quantity measured empirically. In our experiment, this is given by
	\begin{equation*}
	\begin{aligned}
	F_h(\omega_h) &= -2.6 \cdot 10^{-20} \omega^5_h + 4.1 \cdot 10^{-17} \omega^4_h + 3.2 \cdot 10^{-12} \omega^3_h - 7.3 \cdot 10^{-9} \omega^2_h + 2.1 \cdot 10^{-5} \omega_h - 0.0091, \\
	\omega_h(c_h) &= 2.2 \cdot 10^3 c^5_h - 1.7 \cdot 10^2 c^4_h - 4.5 \cdot 10^3 c^3_h - 3 + 10^2 c^2_h + 9.8 \cdot 10^3 c_h - 9.2.
	\end{aligned}
	\end{equation*}
	
	\item $M_{h2}$ is the moment of friction depending on the angular velocity of the beam vertically, given by
	\begin{equation*}
	M_{h2} = - \dot{\alpha_h} f_h,
	\end{equation*}
	where $f_h$ is a constant.
	
	\item $M_{h3}$ is the cross moment from vertical control input, given by
	\begin{equation*}
	M_{h3} = c_v k_{vh},
	\end{equation*}
	where $k_{vh}$ is a constant.
	
	\item $M_{h4}$ is the damping torque from rotating propeller, given by
	\begin{equation*}
	M_{h4} = - a_2 \dot{\alpha_h} \abs(\omega_h),
	\end{equation*}
	where $a_2$ is a constant.
	\end{enumerate}
	
	\item $J_h$ the components of the moment of inertia relative to the vertical axis depends on the vertical position of the beam, given by
	\begin{equation*}
	\begin{aligned}
	J_h &= \frac{m_m}{3}(l_m \cos(\alpha_v))^2 + \frac{m_t}{3} (l_t \cos(\alpha_v))^2 + \frac{m_b}{3} (l_b \sin(\alpha_v))^2 + m_{tr} (l_t \cos(\alpha_v))^2 + m_{mr} (l_m \cos(\alpha_v))^2 \\
	& + m_{cb} (l_{cb} \sin(\alpha_v))^2 + \frac{m_{ts}}{2} r_{ts}^2 + m_{ts}(l_t \cos(\alpha_v))^2 + m_{ms}r^2_{ms} + m_{ms}(l_m \cos(\alpha_v))^2.
	\end{aligned}
	\end{equation*}
\end{itemize}

It appears to have many parameters but all of them are predetermined and the only independent variables are the two voltages applied, namely $c_v$ and $c_h$. We try to learn the model by showing that an optimised selection of the voltages is better than randomly selected ones.

\subsubsection{Hyperparameters settings}

Table \ref{hyp-table} summarises key hyperparameters used for the experiments. For fairness, we use these settings for all random exploration, \emph{latest} mutual information (latMI) strategy and \emph{total} mutual information (totMI) strategy.
\begin{table}[h]
\caption{Key hyperparameters for experiments. }
\label{hyp-table}
\vskip 0.15in
\begin{center}
\begin{small}
\begin{tabular}{lccr}
Experiment & kernel & parameter initials & Training epochs \\
Simulated function & Squared exponential & $\sigma^2 = 1.0, \lambda = 1.0$ & 100 \\
Pendulum & Mat\'ern 32 & $\sigma^2 = 0.3, \lambda = 5.0$ & 200 \\
Cart-Pole & Mat\'ern 32 & $\sigma^2 = 1.0, \lambda = 1.0$ & 200 \\
TRAS & Squared exponential & $\sigma^2 = 1.0, \lambda = 1.0$ & 500 \\
\end{tabular}
\end{small}
\end{center}
\vskip -0.1in
\end{table}

\subsection{Further Results}

\subsubsection{Running Time Analysis}
\begin{figure}[h!]
\centering
\includegraphics[width=.4\linewidth]{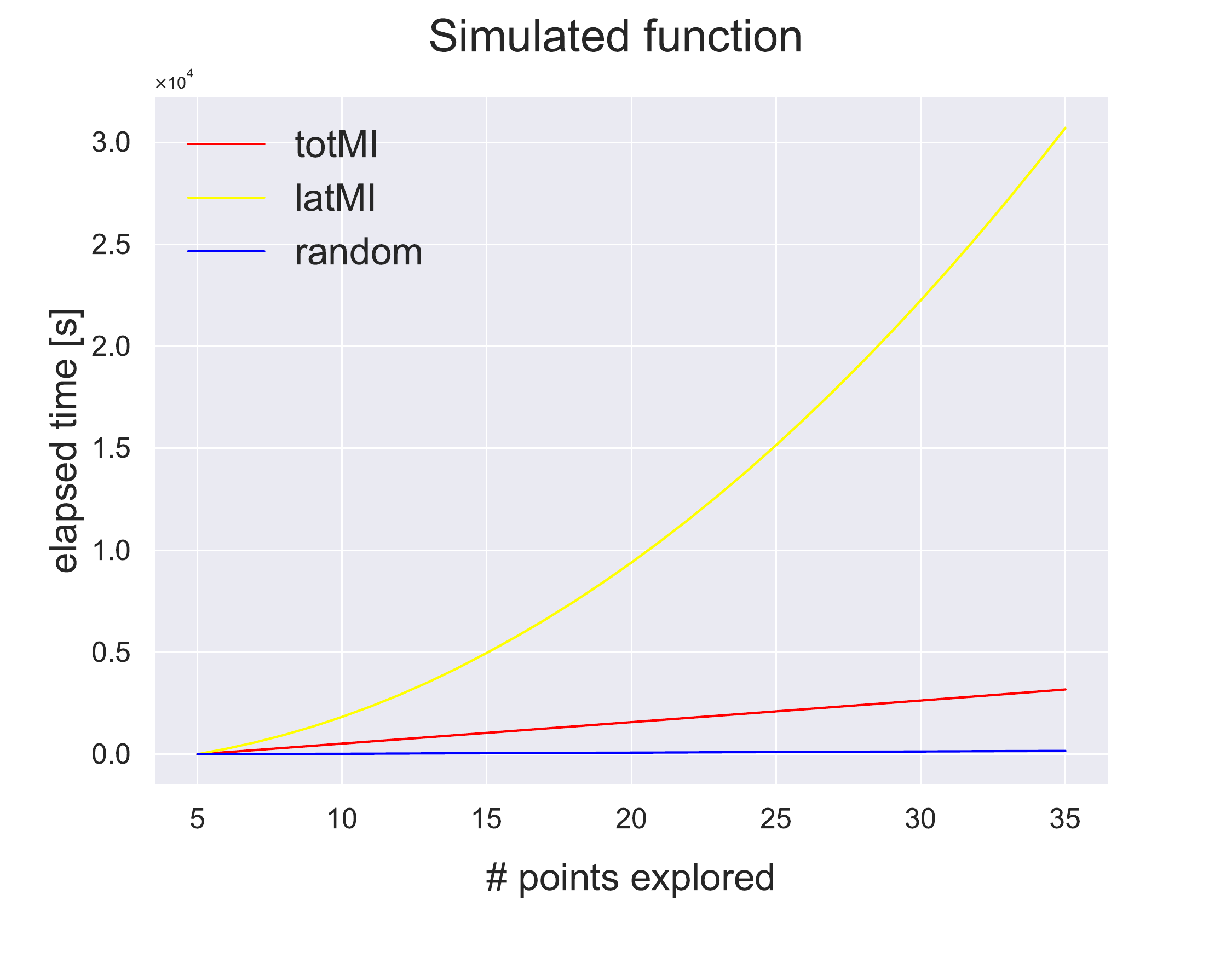}
\includegraphics[width=.4\linewidth]{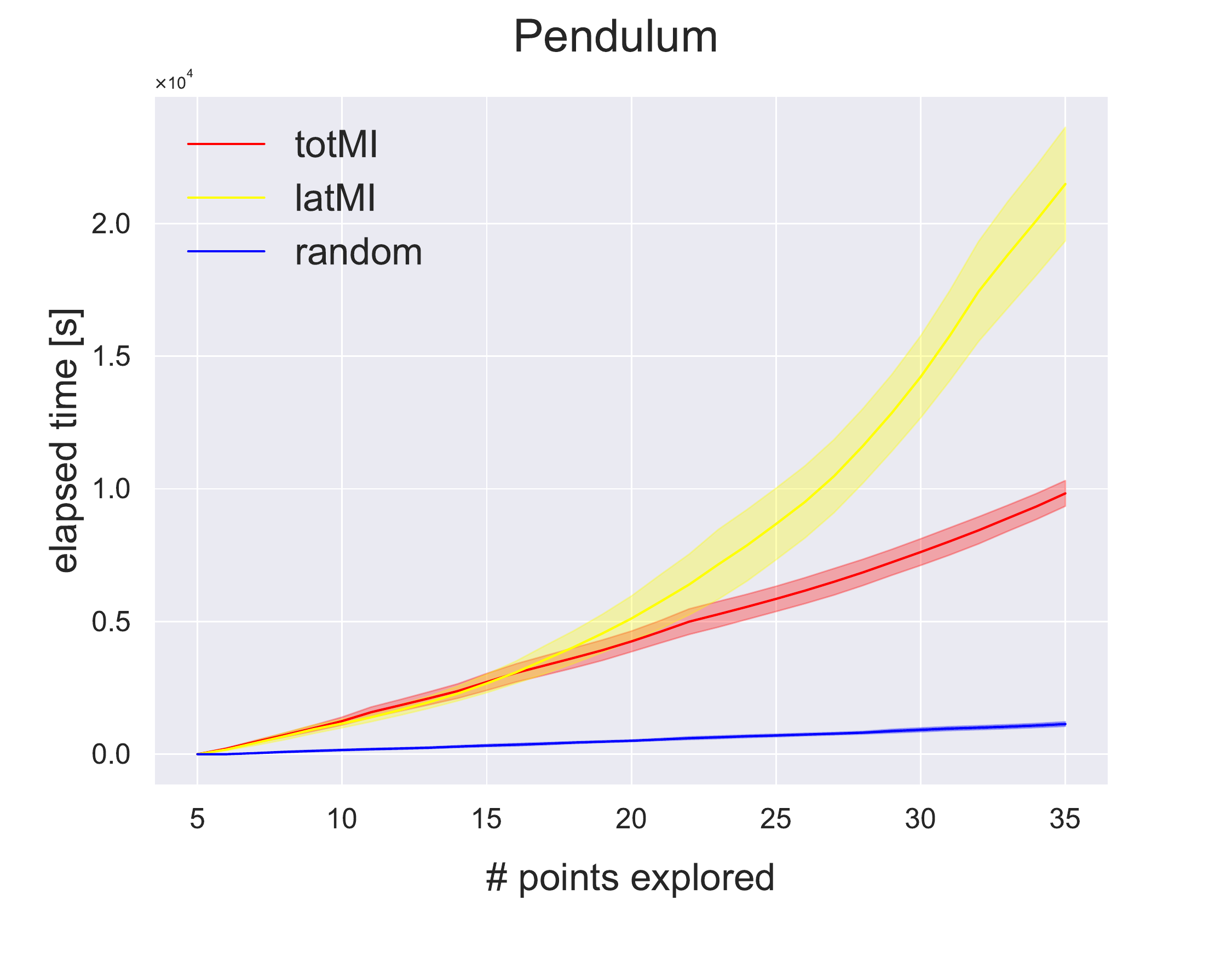}
\includegraphics[width=.4\linewidth]{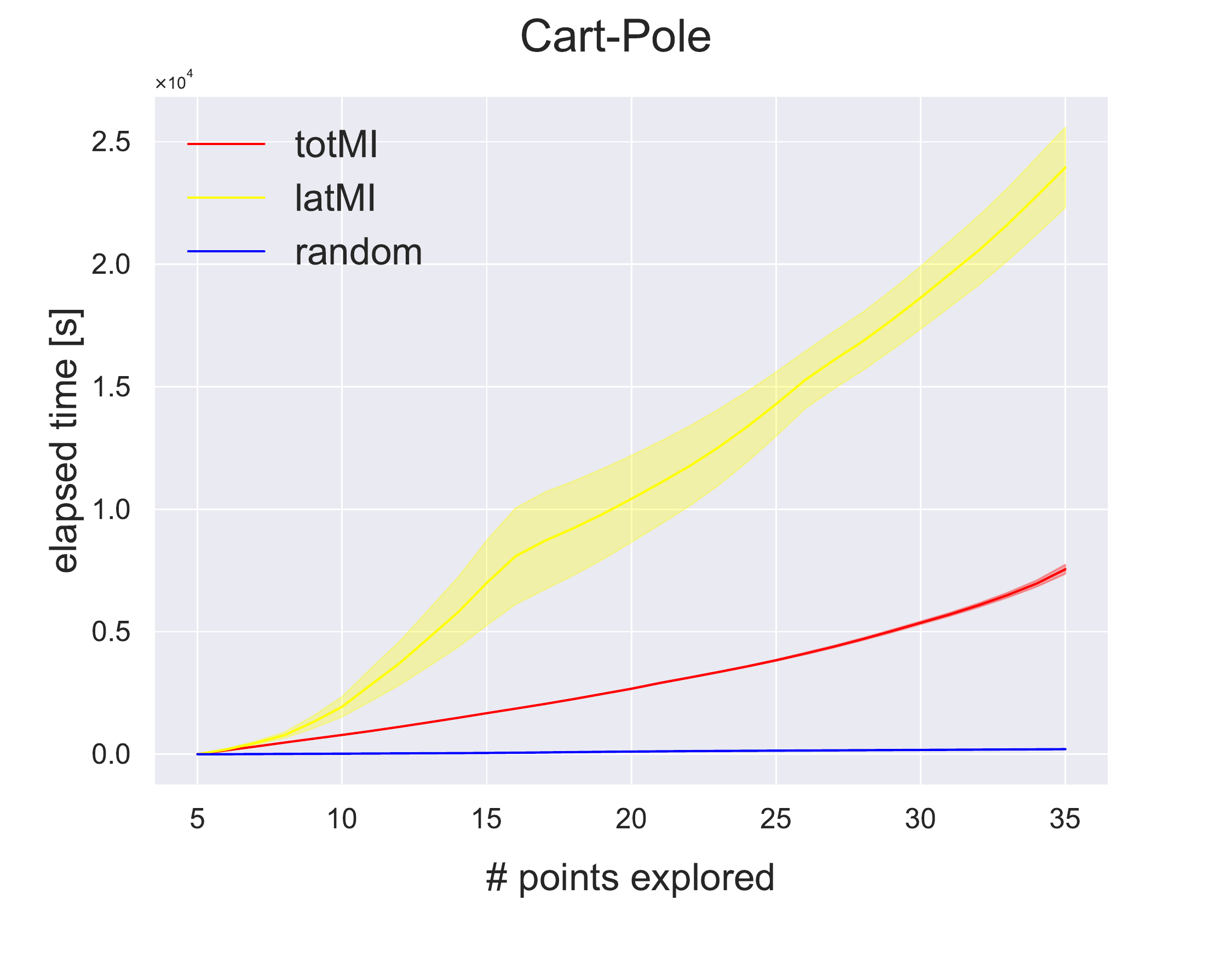}
\includegraphics[width=.4\linewidth]{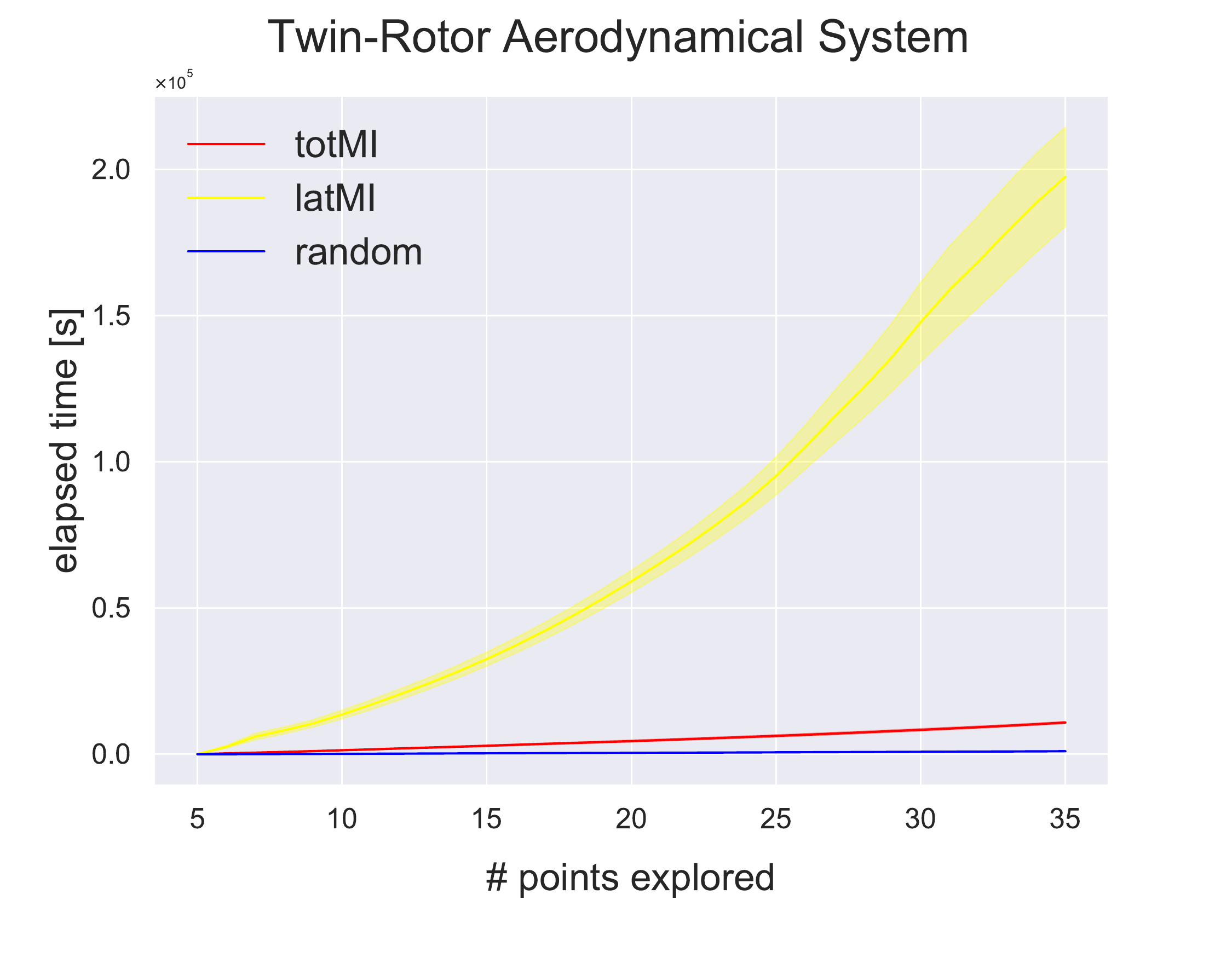}
\caption{Elapsed time for the exploration under simulated function (upper left), pendulum (upper right), cart-pole (lower left) and Twin-Rotor Aerodynamical System (lower right). latMI is presented in yellow; totMI in red and random exploration in blue.} 
\label{fig_time}
\end{figure}

In addition to the predictive performance, we also present the exploration time depending the number of points. The elapsed time is displayed in Figure \ref{fig_time}. While it is impossible to beat random exploration because there is no need to optimise for the control, we show that the totMI strategy runs significantly faster than the latMI strategy. This is because we leverage from the inference scheme for the calculation of $p(\bm{y}_{1:t})$, which saves a significant amount of computational effort. This effect applies to all four experiments we investigate in this paper.

\subsubsection{Comparison with Capone et al.\cite{capone2020localized}}

Due to the lack of other benchmarks on the setting of unobservable states, we also investigate the performance of our algorithm  in a scenario with observable states. Here, we can benchmark against \cite{capone2020localized}. Note that such comparison is by nature not fair. First, our approach assumes latent states whereas their approach requires that states are observable. Second, by having the value of states, GPSSM can be trained in a similar fashion as in GP regression and they resolve this two optimisation problems to be solved in parallel. Therefore, the number of hyperparameters from their settings is rather different from ours. To this end, we keep their initial hyperparameter settings for simplicity. 

Some settings from \cite{capone2020localized} are adjusted to be the same as our settings to facilitate a more plausible comparison, including the number of training epochs, taking arithmetic mean for different trials, and the number of points explored before the model is updated.

We choose Pendulum and Cart-Pole to run the `comparison'. This is because the true value of the angular velocity in these two problems can be computed. Since we have shown from the previous subsection that in addition to the inferior performance, running time for latMI was prohibitively high in comparison to totMI. Therefore, we only look into totMI strategy. We focus on random exploration, totMI and the approach from \cite{capone2020localized}, denoted by \emph{locAL}.

\begin{figure}[h!]
\centering
\includegraphics[width=.4\linewidth]{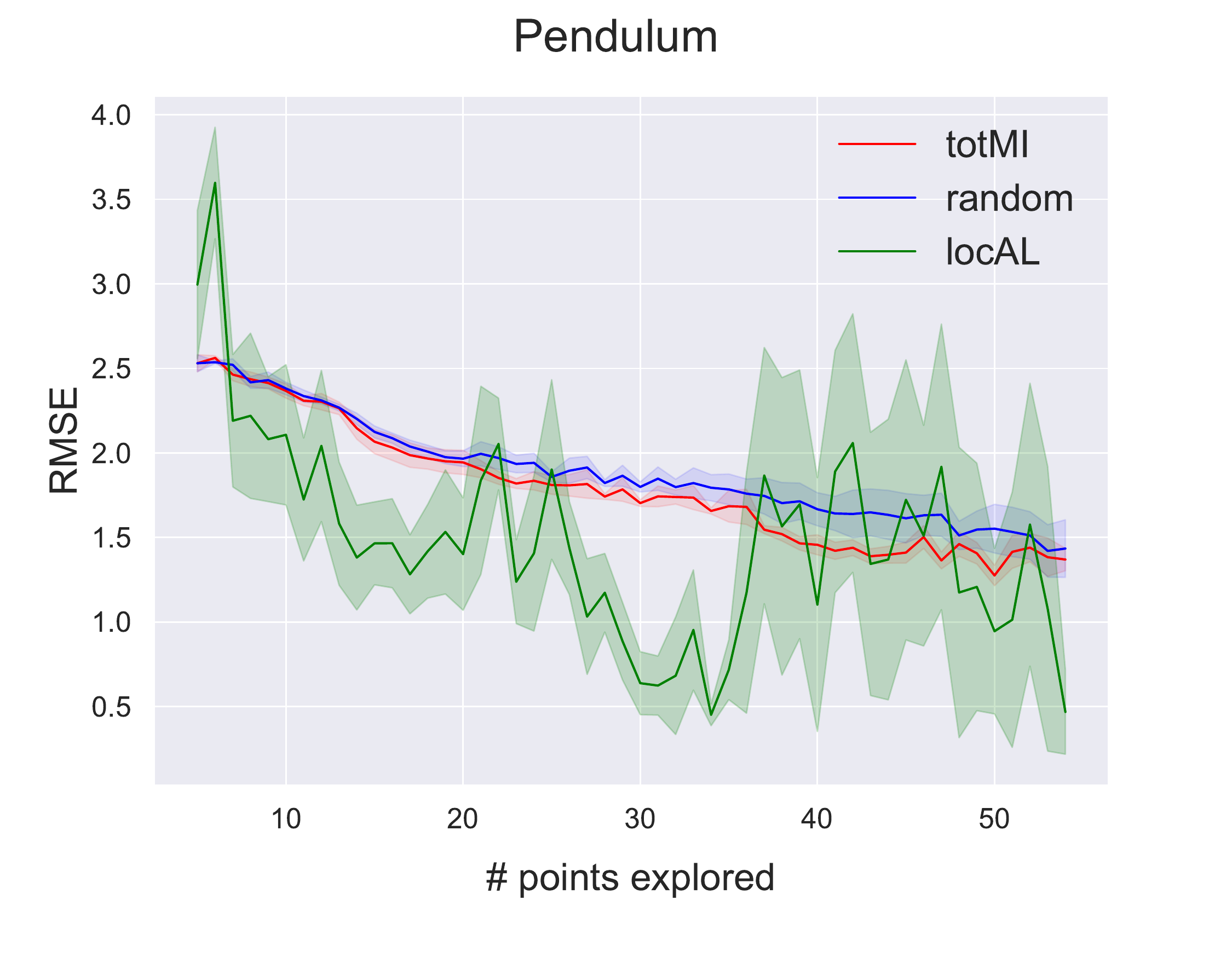}
\includegraphics[width=.4\linewidth]{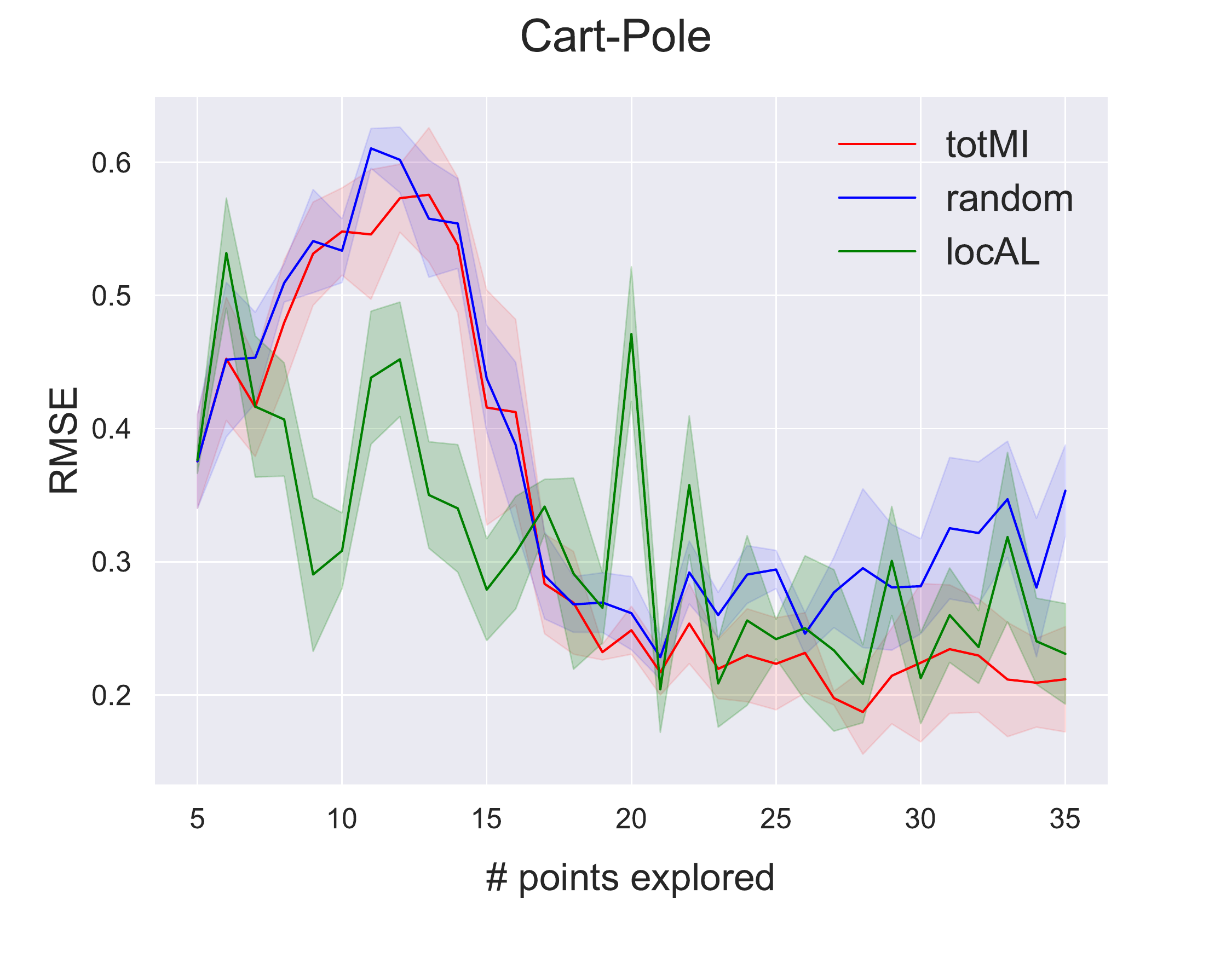}
\caption{Comparison with \cite{capone2020localized} on Pendulum (left) and Cart-Pole (right). totMI in red, random exploration in blue and the approach from \cite{capone2020localized}, denoted by locAL, in green.} 
\label{fig_compCapone}
\end{figure}

Results are summarised in Figure \ref{fig_compCapone}. locAL showed a very unstable improvement in RMSE. Specifically, for pendulum, their results appear to be better in average, but not significant. Whereas for cart-pole, the RMSE begins to decay faster. This could be explained by the fact that their approach collected the information of the states after each round of exploration, which allows the model to be stable quicker. However, by the time when all approaches becomes stable, our totMI strategy still demonstrates a stronger performance. Our approach processes a more stable improvement in RMSE and a competitive performance, compared to the approach by \cite{capone2020localized} which requires the knowledge of states.

\subsection{Further discussion between GPSSM and Reinforcement Learning}
We would also like to address that the strategy of maximising mutual information has also been used in the reinforcement learning (RL) community, especially in planning and control. For instance, Ding et al.\cite{ding2020mutual} applied such strategy in Partially Observable Markov Decision Process (POMDP) framework in order to efficiently learn the RL model. However, there are key differences between AL in GPSSM and efficient learning in POMDP. First, the word \emph{control} is interpreted very differently. In GPSSM, this refers to an augmented latent state and each input is simply a real number (or a real-valued vector if we have multiple controls per time step). This quantity is always deterministic because it is a user-input to steer the exploration of the model. On the other hand, in POMDP, this is a mapping between an action taken in certain state to maximise the expected return or reward. \emph{Optimising control} in this context is essentially finding the optimal policy in RL. Second, AL in GPSSM is a very specific problem formulation and this paper is presented to tackle this whereas efficient learning in POMDP is a general framework with a large room of freedom to specify different components, which leads to various research topics. Afterall, AL for SSM with GPs and RL are two different subfields in machine learning despite GPs are often applied as a tool in RL to help achieve their objective (e.g. \cite{engel2005reinforcement, grande2014sample, berkenkamp2017safe}).

